\documentclass[runningheads]{llncs}
\usepackage{longtable}
\usepackage{pgf,pgfmath,tikz}
\usepackage{subcaption}
\usepackage{pgfplots}

\pgfplotsset{compat=1.18}
\usepackage{sansmath}
\usetikzlibrary{patterns}
\usepackage{caption} 
\usepackage{wrapfig}
\usepackage{annotate-equations}

\usepackage[T1]{fontenc}
\usepackage[T5]{fontenc}
\usepackage{graphicx}
\usepackage{amsmath}
\usepackage{amssymb} 
\usepackage{graphicx}
\usepackage{booktabs}
\usepackage{pifont}
\renewcommand{\checkmark}{\ding{51}}%
\newcommand{\xmark}{\ding{55}}%

\usepackage{adjustbox} 
\usepackage{array} 
\usepackage{arydshln}
\usepackage{arydshln}
\usepackage{algorithm}  
\usepackage{bbding}
\usepackage{algpseudocode}

\usepackage{hyperref}
\usepackage{multirow}

\usepackage{esvect}
\usepackage{orcidlink}

\definecolor{BrickRed}{cmyk}{0,0.89,0.94,0.28}
\definecolor{OliveGreen}{cmyk}{0.64,0,0.95,0.40}
\definecolor{WildStrawberry}{cmyk}{0,0.96,0.39,0}
\definecolor{Plum}{cmyk}{0.50,1,0,0}
\definecolor{NavyBlue}{cmyk}{0.94,0.54,0,0}
\definecolor{Cyan}{cmyk}{1,0,0,0}

\begin{document}
%

\title{A Benchmark Dataset and Evaluation Framework for Vietnamese Large Language Models in Customer Support}
\titlerunning{CSConDa}
%
\author{Long S. T. Nguyen\orcidlink{0009-0008-7488-4714} \and Truong P. Hua \and Thanh M. Nguyen \and Toan Q. Pham \and Nam K. Ngo \and An X. Nguyen \and Nghi D. M. Pham \and Nghia H. Nguyen\and
Tho T. Quan\thanks{Corresponding author}\orcidlink{0000-0003-0467-6254}}
\authorrunning{Long S. T. Nguyen et al.}
%
\institute{URA Research Group, Faculty of Computer Science and Engineering, Ho Chi Minh City University of Technology (HCMUT), Ho Chi Minh City, Vietnam \and Vietnam National University Ho Chi Minh City,  Ho Chi Minh City, Vietnam }
\maketitle              
\begin{abstract}
With the rapid advancement of Artificial Intelligence, Large Language Models (LLMs) have become indispensable in Question Answering (QA) systems, enhancing response efficiency and reducing human workload, particularly in customer service. The rise of Vietnamese LLMs (ViLLMs) has positioned lightweight open-source models as the preferred choice due to their efficiency, accuracy, and privacy advantages. However, systematic evaluations of their performance in domain-specific contexts remain scarce, making it challenging for enterprises to identify the most suitable LLM for customer support applications, especially given the lack of benchmark datasets reflecting real-world customer interactions. To bridge this gap, we introduce Customer Support Conversations Dataset (CSConDa), a high-quality benchmark comprising over 9,000 QA pairs, meticulously curated from customer interactions with human advisors at a large-scale Vietnamese software company. Covering diverse service-related topics, including pricing inquiries, product availability, and technical troubleshooting, CSConDa serves as a representative dataset for evaluating ViLLMs in real-world scenarios. Furthermore, we present a comprehensive evaluation framework, benchmarking 11 lightweight open-source ViLLMs on CSConDa using not only well-suited automatic metrics but also an in-depth syntactic analysis to uncover their strengths, weaknesses, and underlying linguistic patterns. This analysis provides insights into model behavior, explains performance variations, and identifies critical areas for improvement, guiding future advancements in ViLLM development. Thus, by establishing a robust benchmark for LLM-driven customer service applications, our work provides a quantitative evaluation dataset and a comprehensive ViLLM performance comparison, offering key insights into intrinsic model performance, including accuracy, fluency, and consistency, while enabling informed decision-making for next-generation QA systems. Our dataset is publicly available on \href{https://huggingface.co/datasets/ura-hcmut/Vietnamese-Customer-Support-QA}{Hugging Face}.


\end{abstract}

\section{Introduction}
The rapid advancement of \textit{Artificial Intelligence} (AI) has revolutionized \textit{Question Answering} (QA) systems, enabling \textit{Large Language Models} (LLMs) to automate information retrieval and generate responses with varying complexity \cite{Li2024}. As one of the fastest-growing economies, Vietnam is witnessing increasing demand for AI-driven solutions, particularly in customer service, where efficiency and accuracy are paramount. Consequently, many Vietnamese enterprises are integrating LLMs into their QA systems to streamline automated consultation.

\begin{figure}[!ht]
    \centering
    \begin{tikzpicture}[scale=0.7]
    \begin{axis}[
        width=\textwidth,
        height=0.5\textwidth,
        ybar=0pt,
        bar width=8pt,
        xlabel={\small Model Size Category},
        ylabel={\small Number of Models},
        xlabel style={yshift=8pt},
        ylabel style={yshift=-1pt},
        symbolic x coords={
            {0-1B},{1-2B},{2-3B},{3-5B},{5-7B},{7-9B},
            {9-11B},{11-13B},{13-15B},{15-20B},{20-50B},
            {50-100B},{100-500B},{>500B}
        },
        xtick=data,
        x tick label style={font=\scriptsize, rotate=45, anchor=east},
        y tick label style={font=\scriptsize},
        ymin=0,
        ytick={0,200,400,600,800,1000,1200},
        enlarge x limits=0.05,
        legend style={
            at={(0.85,0.98)},
            anchor=north,
            legend columns=1,
            font=\scriptsize
        }
    ]
        \addplot[fill=NavyBlue, draw=NavyBlue!150] coordinates {
            (0-1B,124)
        (1-2B,145)
        (2-3B,83)
        (3-5B,142)
        (5-7B,101)
        (7-9B,1075)
        (9-11B,86)
        (11-13B,71)
        (13-15B,371)
        (15-20B,49)
        (20-50B,297)
        (50-100B,269)
        (100-500B,78)
        (>500B,1)
        };
        \addplot[fill=BrickRed, draw=BrickRed!150] coordinates {
            (0-1B,151)
        (1-2B,164)
        (2-3B,82)
        (3-5B,166)
        (5-7B,67)
        (7-9B,1290)
        (9-11B,174)
        (11-13B,86)
        (13-15B,222)
        (15-20B,50)
        (20-50B,184)
        (50-100B,199)
        (100-500B,60)
        (>500B,3)
        };
        \legend{Top Likes, Trending}
    \end{axis}
\end{tikzpicture}
    \caption{Distribution of LLM model sizes based on popularity metrics on HF.}
    \label{fig:trend}
\end{figure}
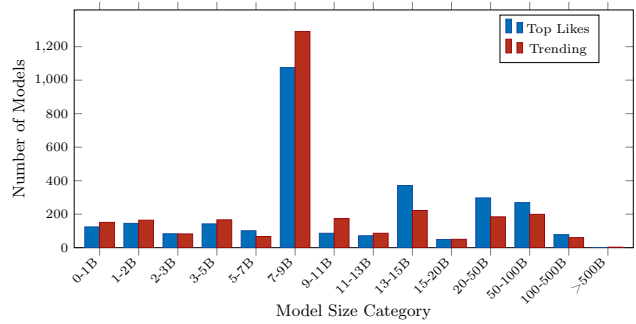

Model size plays a crucial role in LLM performance, with larger models generally exhibiting superior linguistic capabilities \cite{truong-etal-2024-crossing}. While proprietary closed-source models, such as GPT-4\footnote{\url{https://openai.com/index/gpt-4/}}, dominate large-scale applications, open-source alternatives offer greater adaptability for domain-specific tasks, scalability, and enhanced data privacy. Our analysis of model usage statistics on \textit{Hugging Face\footnote{\url{https://huggingface.co/}}} (HF), a leading platform for hosting and fine-tuning open-source LLMs, reveals that as of January 2025, among the top 3,000 most frequently downloaded and widely adopted models, those within the 7–9 billion parameter range are the most preferred, as illustrated in Figure~\ref{fig:trend}. This trend underscores the growing preference for \textit{lightweight open-source models}, which balance computational efficiency and practical usability \cite{wan2024efficient}, making them particularly well-suited for enterprises developing LLM-based QA systems.

Despite the increasing availability of lightweight LLMs in general and \textit{Vietnamese LLMs} (ViLLMs) in particular, enterprises developing LLM-based customer support systems face challenges in selecting the most suitable model for their needs. Therefore, a comprehensive evaluation of LLMs in domain-specific QA is crucial to ensuring their reliability, scalability, and hallucination tendencies before applying external knowledge augmentation or other enhancements. As a result, a benchmark dataset tailored for customer support is essential for systematically comparing ViLLM performance. However, existing Vietnamese QA datasets \cite{nguyen-etal-2020-vietnamese,le-etal-2022-vimqa,10.1145/3527631,10.1007/978-3-030-88113-9_44} primarily focus on factual accuracy within structured contexts, often derived from Wikipedia\footnote{\url{https://www.wikipedia.org/}} articles, covering domains such as news, healthcare, and education. Thus, these datasets fail to reflect the nature of real-world customer interactions, where queries frequently include teencode, abbreviations, and domain-specific jargon. This issue is particularly pronounced in non-English languages like Vietnamese, often leading to misinterpretations, unnatural responses, or even complete failure to generate relevant answers. To the best of our knowledge, no existing dataset has been specifically designed for Vietnamese customer service, nor has any study systematically evaluated ViLLMs in specialized domains such as customer support.

To address this gap, we introduce \textit{\textbf{C}ustomer \textbf{S}upport \textbf{Con}versations \textbf{Da}taset} (CSConDa), a large-scale dataset of nearly 10,000 QA pairs extracted from real-world customer interactions with human advisors at DooPage\footnote{\url{https://doopage.com/}}, a Vietnamese software company serving 30,000 customers and 45,000 advisors through its multi-channel support platform. Through rigorous processing and anonymization, CSConDa serves as a comprehensive benchmark for evaluating ViLLMs in customer support tasks. Our dataset covers diverse topics, including pricing, service inquiries, and technical troubleshooting, and is categorized into three types reflecting different levels of complexity. As a representative dataset, CSConDa enables a systematic assessment of lightweight ViLLMs’ intrinsic ability to generate human-like responses without external knowledge augmentation. Using CSConDa, we systematically evaluate 11 widely adopted lightweight ViLLMs within the 7–9 billion parameter range, employing six carefully selected automatic qualitative and quantitative metrics, along with syntactic analysis, to assess accuracy, hallucination tendencies, and fluency in handling practical customer queries. This evaluation highlights each model’s strengths and weaknesses, identifying key areas for improvement. Therefore, our work equips enterprises with a robust dataset that accurately represents real-world customer interactions, reducing the time required for model assessment while providing valuable insights and comparative analyses to support selecting the most suitable ViLLM for practical deployment. Our key contributions are summarized as follows.

\begin{itemize}
\item We conducted a comprehensive survey of publicly available Vietnamese QA datasets, providing an in-depth analysis of their linguistic characteristics.
\item We introduced CSConDa, the first large-scale Vietnamese QA dataset derived from real-world customer service interactions, establishing a foundation for evaluating ViLLMs in customer support.
\item We conducted a systematic evaluation of 11 widely adopted lightweight open-source ViLLMs, assessing their ability to generate human-like responses in domain-specific QA with selected automatic metrics and syntactic analysis.
\item We provided unique and high-quality insights into both the strengths and limitations of existing ViLLMs, identifying key challenges in real-world deployment and offering actionable recommendations to enhance LLM-based customer support systems.
\end{itemize}







\section{Related Works}

\subsection{Human-Generated Vietnamese QA Datasets}
The availability of text-based Vietnamese QA datasets remains highly limited, as Vietnamese is inherently a low-resource language. Existing datasets, such as \cite{nguyen-etal-2020-vietnamese,le-etal-2022-vimqa}, construct QA pairs from Wikipedia articles, covering various domains. Meanwhile, domain-specific QA datasets, such as \cite{10.1145/3527631,10.1007/978-3-030-88113-9_44}, contain open-ended questions related to news, healthcare, and education, primarily compiled from online news sources and academic materials. Additionally, datasets like \cite{9247161,dao2023vnhsge} focus on multiple-choice questions designed for Vietnamese educational assessments.  While these datasets provide valuable linguistic resources, they are not well-suited for evaluating ViLLMs in customer support applications, where real-world queries are inherently unstructured and often incorporate teencode, abbreviations, and code-switching between Vietnamese and English. More critically, they lack the conversational spontaneity and domain-specific complexity found in real-world customer support interactions.  To bridge this gap, we introduce the first large-scale dataset specifically designed for customer support QA, establishing a representative benchmark for evaluating ViLLMs in practical applications. Additionally, we conduct a comprehensive survey of publicly available text-based human-generated Vietnamese QA datasets, analyzing their statistical properties and linguistic structures in detail, as discussed in Section \ref{subsub:comparedatasets}.

\subsection{Comprehensive Evaluation of ViLLMs}
Assessing ViLLMs in QA tasks presents significant challenges due to the scarcity of high-quality datasets, making domain-specific evaluation even more complex. As a result, few studies have systematically analyzed the performance of ViLLMs, particularly in specialized domains.  To the best of our knowledge, the only existing study related to our work is \cite{truong-etal-2024-crossing}, presented at the flagship conference \textit{North American Chapter of the Association for Computational Linguistics} (NAACL 2024). This study provides a pioneering evaluation of ViLLMs across various NLP tasks using an extensive set of metrics, including QA, on widely adopted Vietnamese datasets. However, these datasets primarily consist of structured text and lack the spontaneity, informality, and domain-specific nuances inherent to customer support conversations. Furthermore, this study does not focus on domain-specific QA and evaluates a broad range of ViLLMs, including both open-source and closed-source models, without addressing their applicability to real-world customer service scenarios.  In contrast, we present the first large-scale evaluation of 11 open-source ViLLMs, specifically targeting the most widely adopted model segment (7–9 billion parameters) and focusing on domain-specific QA in customer support applications. Our findings provide valuable insights into model capabilities, limitations, and areas for improvement, bridging the gap between academic benchmarks and real-world deployment.


\section{The CSConDa Benchmark Dataset}

\subsection{Dataset Creation}
\begin{figure}
    \centering
    \includegraphics[width=\linewidth]{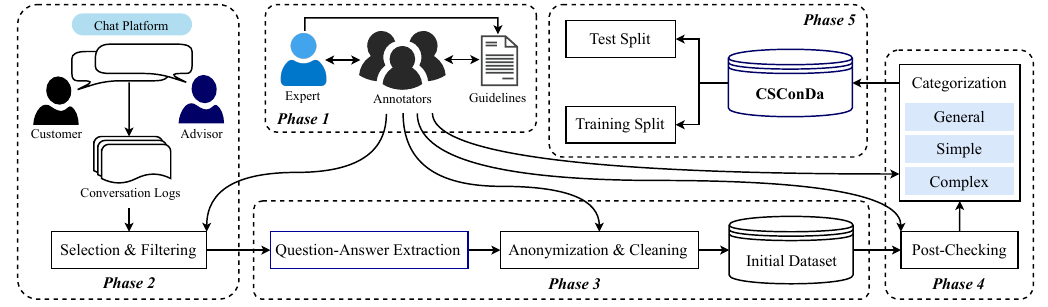}
    \caption{The five-phase process of CSConDa dataset creation.}
    \label{fig:create}
\end{figure}

CSConDa was constructed through five phases: \textit{(i) worker recruitment}, \textit{(ii) conversation collection}, \textit{(iii) dataset creation}, \textit{(iv) validation and categorization}, and \textit{(v) dataset splitting}, as illustrated in Figure~\ref{fig:create}.

\noindent \textbf{Phase 1: Worker Recruitment} \quad A team of ten \textit{annotators}, supervised by a domain \textit{expert} in customer service, was recruited. The expert, also a Chief Executive Officer in the business sector, established \textit{guidelines} for conversation selection, anonymization, and dataset categorization. All annotators signed confidentiality agreements, underwent structured training, and maintained continuous communication with the expert.

\medskip
\noindent \textbf{Phase 2: Conversation Collection} \quad Raw \textit{conversation logs} were sourced from a \textit{chat platform}, where \textit{customers} interact with \textit{advisors}. Annotators followed predefined criteria to perform \textit{selection and filtering}, ensuring high-quality exchanges. The filtering process prioritized coherence, topic diversity, and exclusion of sensitive or inappropriate content.

\medskip
\noindent \textbf{Phase 3: Dataset Creation} \quad After filtering, an automated pipeline was developed for \textit{question-answer extraction}, preserving contextual consistency. The extracted QA pairs then underwent \textit{anonymization and cleaning} to remove personally identifiable information and system-generated artifacts (e.g., syntax markers, reaction icons). The output of this phase formed the \textit{initial dataset}.

\medskip
\noindent \textbf{Phase 4: Validation and Categorization} \quad A \textit{post-checking} step ensured compliance with security standards, formatting consistency, and annotation accuracy. The \textit{categorization} process divided the dataset into three types: \textit{General}, \textit{Simple}, and \textit{Complex}, based on conversational complexity, reasoning demands, and required domain knowledge, as defined in the annotation guidelines.

\medskip
\noindent \textbf{Phase 5: Dataset Splitting} \quad After categorizing the dataset, we obtained \textit{CSConDa}, which was then divided into standard \textit{training} and \textit{test} splits. The test split comprises 1,500 representative questions, evenly distributed across the three predefined types, serving as a benchmark for evaluating ViLLMs.

\subsection{Dataset Overview}
CSConDa is available on \href{https://huggingface.co/datasets/ura-hcmut/Vietnamese-Customer-Support-QA}{Hugging Face}, with categorization criteria on the introduction page. Table~\ref{tab:overview} presents examples of its three types, illustrating linguistic variations in practical customer support conversations, including: \textit{(i) \textbf{code-switching}}, frequently in domain-specific terms;
\textit{(ii) \textcolor{NavyBlue}{\textbf{abbreviations}}}, encompassing teencode and informal shorthand;
\textit{(iii) \textcolor{BrickRed}{\textbf{acronyms}}}, formal abbreviations derived from initial letters; and
\textit{(iv) \textcolor{OliveGreen}{\textbf{typos}}}, resulting from fast or imprecise typing.

\begin{table}[!ht]
    \centering
    \renewcommand{\arraystretch}{1.1}
 \caption{Overview of the three predefined types in CSConDa.}
    \label{tab:overview}
    
    \resizebox{\textwidth}{!}{%
    \begin{tabular}{@{}p{15.3cm}@{}} 
        \toprule
        \underline{\textbf{Type:}} General \\
        \underline{\textbf{Question:}}  
        \textcolor{NavyBlue}{\textbf{âu kê}}  
        \textcolor{NavyBlue}{\textbf{thank kiu}}  
        \textcolor{NavyBlue}{\textbf{e}}. (\textbf{English:} Okay, thank you.)\\
        \underline{\textbf{Answer:}}  
        dạ vâng ạ. (\textbf{English:} Yes, sure.)\\
        \midrule
        \underline{\textbf{Type:}} Simple \\
        \underline{\textbf{Question:}}  
        Báo giá giúp \textcolor{NavyBlue}{\textbf{M}} nhé.  
        Gọi \textcolor{NavyBlue}{\textbf{M}} lúc 11\textcolor{BrickRed}{\textbf{h}} số này \textcolor{OliveGreen}{\textbf{nef}} \textless số điện thoại\textgreater.  
        \textcolor{NavyBlue}{\textbf{C}} cần \textcolor{NavyBlue}{\textbf{qly}} 8 \textbf{page}, 2 \textbf{Zalo}, 1 \textbf{website},  
        1 \textcolor{BrickRed}{\textbf{YTB}}, 3-5 người dùng.  
        (\textbf{English:} Please send me the pricing details.  
        Call me at 11 AM at this number \textless phone number\textgreater.  
        I need 8 pages, 2 Zalo accounts, 1 website,  
        and 1 YouTube channel for 3–5 users.)\\ 
        \underline{\textbf{Answer:}}  
        Em gửi chị chi phí cho nhu cầu của mình gồm 8 \textbf{page}, 2 \textbf{Zalo OA}, 1 \textbf{web} (tổng 13 kênh)  
        và 3 hoặc 5 người dùng ạ.  
        Đối với nền tảng \textcolor{BrickRed}{\textbf{YTB}}, hiện tại bên em chưa hỗ trợ.  
        Chi phí trên là giá phần mềm trong 01 tháng.  
        (\textbf{English:} I have sent you the pricing details,  
        including 8 pages, 2 Zalo official accounts (OAs), 1 website (total 13 channels),  
        and 3 or 5 users. Currently, we do not support YouTube.  
        The above price covers 1 month of software usage.)\\
        \midrule
        \underline{\textbf{Type:}} Complex \\ 
        \underline{\textbf{Question:}}  
        à thế đây là \textcolor{OliveGreen}{\textbf{teen}} \textbf{maps}. chứ đâu phải tên \textbf{business} đâu em.  
        tên 1 địa chỉ \textbf{map} đó.  
        Nhưng \textcolor{NavyBlue}{\textbf{lsao}} mà gõ tên \textbf{map} vào đó được.  
        trong khi bên trong cho phép \textbf{add} nhiều \textbf{locations}?  
        Lỗi \textbf{file} này \textcolor{NavyBlue}{\textbf{e}} ơi,  
        \textcolor{NavyBlue}{\textbf{ko}} \textbf{down} \textcolor{NavyBlue}{\textbf{đc}}.  
        Bên \textbf{Zalo OA} \textbf{down} \textcolor{NavyBlue}{\textbf{bt}}.  
        (\textbf{English:} Oh, so this is the name on Google Maps,  
        not the business name.  
        This is just a map location.  
        But how can I enter a map name when multiple locations are allowed?  
        This file is corrupted and cannot be downloaded.  
        On Zalo OA, the download works fine.)\\
        \underline{\textbf{Answer:}}  
        dạ phần tên này bên \textbf{Google} họ để trong \textbf{API} là tên liên quan \textbf{Google Business} ạ.  
        Phần này chắc để em \textbf{check} lại thêm với \textbf{dev} ạ.  
        File \textbf{OA} để em báo \textbf{dev} kiểm tra lại ạ.  
        Anh ơi, anh có thể cho em xin \textbf{link file Excel} bị lỗi ở trên \textbf{Zalo OA} được không ạ?  
        (\textbf{English:} Yes, Google categorizes this name under  
        Google Business in their application programming interface (API).  
        I will verify this with our development team.  
        Regarding the file issue, I will report it to the developers.  
        Could you send me the link to the corrupted Excel file via Zalo OA?)\\
        \bottomrule
    \end{tabular}
    }
\end{table}

\vspace*{-0.8cm}
\subsection{Dataset Analysis}

\begin{wraptable}{r}{0.5\textwidth}
\vspace*{-1.1cm}
    \centering
    \caption{Overall summary of CSConDa.}
    \renewcommand{\arraystretch}{1.1}
    \setlength{\tabcolsep}{3pt}
    \resizebox{0.45\textwidth}{!}{%
    \begin{tabular}{lccc}
        \toprule
        & \textbf{Training} & \textbf{Test} & \textbf{All} \\
        \midrule
        Number of samples & 8,349 & 1,500 & 9,849 \\
        Question length  & 16.31 & 19.69 & 16.82 \\
        Answer length  & 41.37 & 29.75 & 39.60 \\
        Vocabulary size & 4,432 & 2,274 & 4,683 \\
        Abbreviation count & 10,164 & 2,179 & 12,343 \\
        Acronym count & 5,710 & 1,191 & 6,901 \\
        Typos count & 1,537 & 270 & 1,807 \\
        Abbreviation frequency & 0.10 & 0.09 & 0.10 \\
        Acronym frequency & 0.06 & 0.05 & 0.07 \\
        Typos frequency & 0.02 & 0.01 & 0.02 \\
        \bottomrule
    \end{tabular}%
    }
    \label{tab:dataset_total_stats}
    \vspace*{-0.5cm}
\end{wraptable}
\textbf{Overall Statistics} \quad Table~\ref{tab:dataset_total_stats} presents key statistics on the structural composition of CSConDa, detailing the occurrence and frequency of acronyms, abbreviations, and typos within sentences. These features are first averaged at the sentence level, then across all records in each split to obtain the final values. Notably, customer questions exhibit a high prevalence of informal linguistic patterns, such as teencode and shorthand expressions, while answers are consistently longer. These trends reflect common characteristics of customer support interactions.

\medskip \noindent \textbf{Type-based Statistics} \quad Table~\ref{tab:dataset_detailed_stats} presents a detailed breakdown of CSConDa statistics across different types in each split. A notable observation is that average question length increases progressively with type complexity. The test split, which serves as the benchmark dataset for evaluation, is evenly distributed across the three types, ensuring a diverse representation of linguistic characteristics.

\begin{table}[!ht]
    \centering
    \caption{Detailed analysis of CSConDa across different types.}
    \renewcommand{\arraystretch}{1.1}
    \setlength{\tabcolsep}{3pt}
    \resizebox{\textwidth}{!}{%
    \begin{tabular}{lccccccccc}
        \toprule
        & \multicolumn{3}{c}{\textbf{Training}} & \multicolumn{3}{c}{\textbf{Test}} & \multicolumn{3}{c}{\textbf{All}} \\
        \cmidrule(lr){2-4} \cmidrule(lr){5-7} \cmidrule(lr){8-10}
        & \textbf{General} & \textbf{Simple} & \textbf{Complex}
        & \textbf{General} & \textbf{Simple} & \textbf{Complex}
        & \textbf{General} & \textbf{Simple} & \textbf{Complex} \\
        \midrule
        Number of samples & 3,023 & 4,711 & 614 & 500 & 500 & 500 & 3,523 & 5,211 & 1,114 \\
        Question length& 9.01 & 18.03 & 38.95 & 10.16 & 20.11 & 28.79 & 9.18 & 18.23 & 34.39 \\
        Answer length & 43.24 & 39.21 & 48.70 & 14.41 & 29.34 & 45.50 & 39.16 & 38.27 & 47.26 \\
        Vocabulary size & 2,497 & 3,678 & 2,072 & 964 & 1,446 & 1,663 & 2,624 & 3,782 & 2,445 \\
        Abbreviation count & 2,208 & 6,758 & 1,198 & 434 & 852 & 893 & 2,642 & 7,610 & 2,091 \\
        Acronym count & 1,339 & 3,739 & 632 & 263 & 456 & 472 & 1,602 & 4,195 & 1,104 \\
        Typos count & 235 & 1,133 & 169 & 58 & 104 & 108 & 293 & 1,237 & 277 \\
        Abbreviation frequency & 0.10 & 0.10 & 0.08 & 0.10 & 0.10 & 0.08 & 0.10 & 0.10 & 0.08 \\
        Acronym frequency & 0.08 & 0.05 & 0.04 & 0.06 & 0.06 & 0.04 & 0.06 & 0.05 & 0.04 \\
        Typos frequency & 0.01 & 0.02 & 0.01 & 0.01 & 0.01 & 0.01 & 0.01 & 0.12 & 0.01 \\

        \bottomrule
    \end{tabular}%
    }
    \label{tab:dataset_detailed_stats}
\end{table}

\noindent \textbf{Comparison with Existing Datasets} \quad
\label{subsub:comparedatasets}
We conduct a comprehensive comparison of publicly available Vietnamese QA datasets, focusing exclusively on text-based QA. Our analysis examines key characteristics, including the \textit{number of samples} (NoS), \textit{average question length} (AQL), \textit{average answer length} (AAL), and the \textit{presence of abbreviations} (Abb.), \textit{acronyms} (Acr.), and \textit{typos} (Typ.), as summarized in Table~\ref{tab:qa_datasets}. Unlike prior datasets, which are primarily derived from structured or semi-structured sources such as Wikipedia, online articles, and educational materials, CSConDa is uniquely constructed from real-world human interactions. As a result, it is the only dataset that authentically captures the informal and context-dependent nature of customer support conversations. Another key distinction lies in the QA paradigm, while most existing Vietnamese QA datasets adopt an extractive approach, where answers are directly retrieved from a predefined context, CSConDa aligns more closely with conversational QA, where responses are inherently open-ended and require contextual reasoning.

\begin{table*}[!ht]
    \centering
    \caption{Comparison of CSConDA with existing Vietnamese QA datasets.}
    \label{tab:qa_datasets}
    \renewcommand{\arraystretch}{1.1}
    \setlength{\tabcolsep}{3pt} 
    \begin{adjustbox}{width=\textwidth}
    \begin{tabular}{l p{2.5cm} l c c c c c c p{4.25cm}}
        \toprule
        \textbf{Dataset} & \textbf{QA Task} & \textbf{Domain} & \textbf{NoS} & \textbf{AQL} & \textbf{AAL} & \textbf{Abb.} & \textbf{Acr.} & \textbf{Typ.} & \textbf{Source} \\
        \midrule
         \textbf{CSConDA (Ours)} & \textbf{Conversational} & \textbf{Customer Support} & \textbf{9,862} & \textbf{17.2} & \textbf{40.7} & \checkmark & \checkmark & \checkmark & \textbf{Human Conversations} \\
        UIT-ViQuAD \cite{nguyen-etal-2020-vietnamese}  & Extractive & General & 23,074 & 12.2 & 8.2 & \xmark & \checkmark & \xmark & Wikipedia \\
        UIT-ViCoQA \cite{10.1007/978-3-030-88113-9_44} & Conversational & General & 10,000 & 9.4 & 9.7 & \xmark & \checkmark & \xmark & Online Healthcare Articles \\
        UIT-ViWikiQA \cite{10.1007/978-3-030-82147-0_42}  & Extractive & General & 23,074 & 14.49	& 39.21 & \xmark & \checkmark & \xmark & Wikipedia \\ 
        VIMQA \cite{le-etal-2022-vimqa} & Extractive, Multi-Hop & General & 9,044 & 15.9 & 2.7 & \xmark & \checkmark & \xmark & Wikipedia \\
        VnYQA \cite{https://doi.org/10.1155/2021/6550871} & Extractive & General & 100 & 14.0 & 10.0 & \xmark & \checkmark & \xmark & Wikipedia  \\
        UIT-ViMMRC \cite{9247161} & Multiple-Choice & General & 2,783 & 12.5 & 7.5 & \xmark & \checkmark & \xmark & Primary School Reading Texts \\
        ViMedAQA \cite{tran-etal-2024-vimedaqa} & Extractive, ~~~~~~~~~~~~ Abstractive & Healthcare & 44,313 & 	13.49	& 24.16 & \xmark & \checkmark & \xmark & Online Medical Documents\\
        UIT-ViNewsQA \cite{10.1145/3527631} & Extractive & Healthcare & 22,057 & 10.6 & 10.7 & \xmark & \checkmark & \xmark & Online Healthcare Articles \\
        UIT-ViCoV19QA \cite{thai-etal-2022-uit} & Abstractive, Knowledge-Based & Healthcare & 4,500 & 31.79 & 119.76 & \xmark & \checkmark & \xmark & Official Health Documents \\
        ViHealthQA \cite{10.1007/978-3-031-10986-7_30} & Abstractive, Knowledge-Based & Healthcare & 10,013 & 103.87 & 495.33 & \xmark & \checkmark & \xmark & Online Healthcare Articles \\
        VNHSGE \cite{dao2023vnhsge} & Multiple-Choice & Education & 7,127 & 64.6 & 1.0 & \xmark & \checkmark & \xmark & High School Graduation Tests \\
        ViRHE4QA \cite{do2024r2gqa} & Extractive, ~~~~~~~~~~~~ Abstractive & Education & 9,758 & 17.09 & 24.18 & \xmark & \checkmark & \xmark & University Regulations \\
        \cite{minh2025using} & Knowledge-Based & Education & 985 & 74.3 & 415.6 & \xmark & \checkmark & \xmark & University Regulations \\
        \cite{10.1007/978-3-031-60511-6_4} & Knowledge-Based & Legal & 5,992 & 17.3 & 1.58 & \xmark & \checkmark & \xmark & Civil Law \\
        VlogQA \cite{ngo-etal-2024-vlogqa} & Extractive & Food \& Lifestyle & 8,047 & 10.09 & 3.22 & \xmark & \checkmark & \xmark & YouTube Transcriptions \\
        ViRe4MRC \cite{do-etal-2023-machine} & Extractive & Food \& Lifestyle & 6,637	& 10.72	& 6.37 & \checkmark & \checkmark & \checkmark & Online Food Reviews  \\
        \bottomrule
    \end{tabular}
    \end{adjustbox}
\end{table*}

\section{ViLLMs Comprehensive Evaluation}

\subsection{Proposed Multi-Aspect Evaluation Framework}

\begin{figure}
    \centering
    \includegraphics[width=\linewidth]{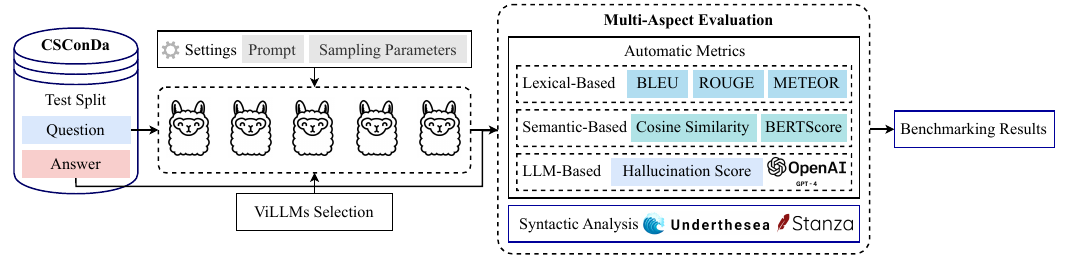}
    \caption{Overview of our multi-aspect evaluation framework for ViLLMs.}
    \label{fig:framework}
\end{figure}

\noindent \textbf{Overall Framework}  \quad  
We present a comprehensive framework for evaluating ViLLMs in domain-specific QA, particularly in customer support, as illustrated in Figure~\ref{fig:framework}. Selected ViLLMs are assessed via inference using questions from the CSConDa test split as input prompts. Our evaluation considers accuracy, fluency, human-likeness, and hallucination tendencies, leveraging automatic metrics benchmarked against human-annotated reference answers. Beyond numerical assessment, we analyze the syntactic structure of ViLLM-generated responses for deeper linguistic insights. These findings identify model strengths and limitations, highlight key areas for improvement, and provide businesses with valuable guidance in selecting and refining ViLLMs.  

\noindent \textbf{ViLLMs Selection}  \quad 
We selected 11 \textit{lightweight open-source ViLLMs} with \textit{state-of-the-art} (SOTA) performance in their respective benchmarks, ranging from 7–9 billion parameters, publicly available as of January 2025. Among them, four models are specifically fine-tuned for Vietnamese: \href{https://huggingface.co/ura-hcmut/ura-llama-2.1-8b}{URA-LLaMa-2.1 8B} \cite{truong-etal-2024-crossing}, \href{https://huggingface.co/ura-hcmut/GemSUra-7B}{GemSUra 7B} \cite{truong-etal-2024-crossing}, \href{https://huggingface.co/Viet-Mistral/Vistral-7B-Chat}{Vistral 7B} \cite{chien2023vistral}, and \href{https://huggingface.co/vilm/vinallama-7b-chat}{VinaLLaMA 7B} \cite{Nguyen2023VinaLLaMALV}. The remaining seven are multilingual models trained on multiple Asian languages, including Vietnamese: \href{https://huggingface.co/SeaLLMs/SeaLLMs-v3-7B-Chat}{SeaLLMs 7B} \cite{nguyen-etal-2024-seallms}, \href{https://huggingface.co/sail/Sailor-7B-Chat}{Sailor 7B} \cite{dou-etal-2024-sailor}, \href{https://huggingface.co/Qwen/Qwen2-7B-Instruct}{Qwen2 7B} \cite{Yang2024Qwen2TR}, \href{https://huggingface.co/ghost-x/ghost-8b-beta-1608}{Ghost 8B} \cite{ghost-8b-beta}, \href{https://huggingface.co/aisingapore/sea-lion-7b-instruct}{SEA-LION 7B} \cite{tjhi-etal-2023-sea}, \href{https://huggingface.co/bigscience/bloomz-7b1}{BLOOMZ 7B} \cite{muennighoff-etal-2023-crosslingual}, and \href{https://huggingface.co/CohereForAI/aya-expanse-8b}{Aya-Expanse 8B} \cite{Dang2024AyaEC}. For benchmarking, we used the latest instruction-tuned or chat-oriented versions to improve domain adaptability and maintain consistent prompt formatting.

\medskip \noindent \textbf{Settings}  \quad  
To ensure fair benchmarking, all ViLLMs were evaluated under identical inference conditions, including \textit{prompting} and \textit{sampling parameters}. Zero-shot prompting was used to assess their intrinsic ability to generate fluent, accurate responses. We fine-tuned sampling parameters to balance output diversity and stability while constraining response length to match the average human-written responses in CSConDa. All experiments were conducted on a single NVIDIA A100 (40GB) GPU to ensure computational uniformity.

\medskip
\noindent
\textbf{Automatic Metrics} \quad  
We employed a \textit{multi-aspect evaluation} framework integrating \textit{lexical-based}, \textit{semantic-based} \cite{chen-etal-2019-evaluating}, and modern \textit{LLM-based} metrics to comprehensively assess ViLLM performance in domain-specific QA. For lexical similarity, we used \textit{BLEU-2} to measure bigram precision, \textit{ROUGE-L} for recall-oriented longest common subsequence overlap, and \textit{METEOR}, which incorporates stemming, synonym matching, and word order flexibility to evaluate word choice, phrasing accuracy, and textual coherence. For semantic alignment, we computed \textit{Cosine Similarity} (Cos. Sim.) at the sentence level and applied \textit{BERTScore} to assess contextual token similarity between ViLLM outputs and human references, both leveraging the SOTA \textit{Vietnamese embedding model}\footnote{\url{https://huggingface.co/dangvantuan/vietnamese-embedding}} to capture deeper semantic relationships beyond surface-level word overlap.  For \textit{hallucination detection} (Hallu. Score), we employed \textit{Kolena's prompt-based metric}\footnote{\url{https://docs.kolena.com/metrics/prompt-based-hallucination-metric/}}, which utilizes \textit{OpenAI's GPT-4} to identify misinformation and fabricated content, ensuring response reliability. However, in real-world scenarios, some ViLLMs fail to generate responses or produce nonsensical loops, significantly impacting usability. To address this, we introduce a \textit{penalty factor} \(\rho\) in the final metric computation. Let \(x_i\) be the individual metric score for the \(i\)-th question, \(A\) the number of successfully generated answers, and \(N\) the total number of test samples. The adjusted score for metric \(M\) is computed as Equation \ref{eq:score}.
\begin{equation}\label{eq:score}
\text{Score}_M = \left(\frac{\sum_{i=1}^{N} x_i}{N}\right) \times \rho, 
\quad 
\rho = \left(\frac{A}{N}\right)^{M_c},
\end{equation}
where \(M_c = 1\) if a higher value of \(M\) indicates better performance and \(M_c = -1\) otherwise. This ensures that models failing to generate valid responses are fairly penalized in the final evaluation.

\begin{figure}[!ht]
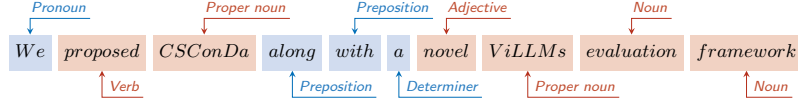

\hspace*{-0.65cm} 
    \makebox[\textwidth]{  
        \resizebox{0.8\textwidth}{!}{  
            \begin{minipage}{\textwidth}
                \centering
                \begin{equation*}
                    \adjustbox{raise=0.5em}{%
                        $\eqnmarkbox[NavyBlue]{pos1}{\strut We}$
                        \hspace{-0.3em}
                        $\eqnmarkbox[BrickRed]{pos2}{\strut proposed}$
                        \hspace{-0.3em}
                        $\eqnmarkbox[BrickRed]{pos3}{\strut CSConDa}$
                        \hspace{-0.3em}
                        $\eqnmarkbox[NavyBlue]{pos4}{\strut along}$
                        \hspace{-0.3em}
                        $\eqnmarkbox[NavyBlue]{pos5}{\strut with}$
                        \hspace{-0.3em}
                        $\eqnmarkbox[NavyBlue]{pos6}{\strut a}$
                        \hspace{-0.3em}
                        $\eqnmarkbox[BrickRed]{pos7}{\strut novel}$
                        \hspace{-0.3em}
                        $\eqnmarkbox[BrickRed]{pos8}{\strut ViLLMs}$
                        \hspace{-0.3em}
                        $\eqnmarkbox[BrickRed]{pos9}{\strut evaluation}$
                        \hspace{-0.3em}
                        $\eqnmarkbox[BrickRed]{pos10}{\strut framework}$
                    }
                \end{equation*}

                \annotate[xshift=-0.3em, yshift=0.8em]{above}{pos1}{\scriptsize \itshape Pronoun}
                \annotate[yshift=-0.3em]{below}{pos2}{\scriptsize \itshape Verb}
                \annotate[xshift=-0.3em, yshift=0.8em]{above}{pos3}{\scriptsize \itshape Proper noun}
                \annotate[yshift=-0.3em]{below}{pos4}{\scriptsize \itshape Preposition}
                \annotate[xshift=-0.3em, yshift=0.8em]{above}{pos5}{\scriptsize \itshape Preposition}
                \annotate[yshift=-0.3em]{below}{pos6}{\scriptsize \itshape Determiner}
                \annotate[xshift=-0.3em, yshift=0.8em]{above}{pos7}{\scriptsize \itshape Adjective}
                \annotate[yshift=-0.3em]{below}{pos8}{\scriptsize \itshape Proper noun}
                \annotate[xshift=-0.3em, yshift=0.8em]{above}{pos9}{\scriptsize \itshape Noun}
                \annotate[yshift=-0.3em]{below}{pos10}{\scriptsize \itshape Noun}
            \end{minipage}
        }
    }
    \vspace*{-0.2cm}
    \caption{Illustration of POS Ratio computation. Content words include nouns, verbs, adjectives, adverbs, and proper nouns, whereas function words consist of pronouns, determiners, and prepositions. The POS ratio for this sentence is computed as \(POS_{C/F} = \frac{7}{3}\).}
\label{fig:postagging}
\end{figure}

\begin{figure}[!ht]
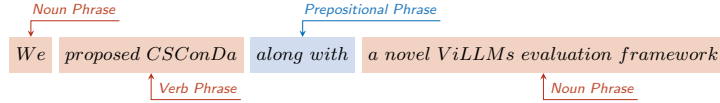

    \centering  
    \makebox[\textwidth]{  
        \resizebox{0.8\textwidth}{!}{  
            \begin{minipage}{\textwidth}
                \centering
                \begin{equation*}
                    \adjustbox{raise=0.5em}{%
                        $\eqnmarkbox[BrickRed]{pos1}{\strut We}$
                        \hspace{-0.3em}
                        $\eqnmarkbox[BrickRed]{pos2}{\strut proposed\ CSConDa}$
                        \hspace{-0.3em}
                        $\eqnmarkbox[NavyBlue]{pos3}{\strut along\ with}$
                        \hspace{-0.3em}
                        $\eqnmarkbox[BrickRed]{pos4}{\strut a\ novel\ ViLLMs\ evaluation\ framework}$
                    }
                \end{equation*}

                \annotate[xshift=-0.3em, yshift=0.8em]{above}{pos1}{\scriptsize \itshape Noun Phrase}
                \annotate[yshift=-0.3em]{below}{pos2}{\scriptsize \itshape Verb Phrase}
                \annotate[xshift=-0.3em, yshift=0.8em]{above}{pos3}{\scriptsize \itshape Prepositional Phrase}
                \annotate[yshift=-0.3em]{below}{pos4}{\scriptsize \itshape Noun Phrase}
            \end{minipage}
        }
    }
    \vspace*{-0.2cm}
    \caption{Illustration of Phrase Ratio computation. Content phrases comprise noun phrases and verb phrases, while function phrases include prepositional phrases. Thus, the phrase ratio for this sentence is determined as \(PH_{C/F} = \frac{3}{1}\).}
\label{fig:phrasetagging}
\end{figure}

\begin{figure}[!ht]
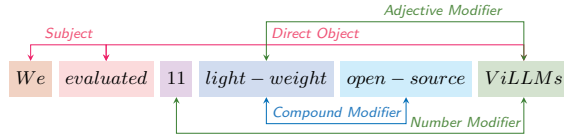

   \centering  
    \makebox[\textwidth]{  
        \resizebox{0.8\textwidth}{!}{  
            \begin{minipage}{\textwidth}
                \centering
                \begin{equation*}
                    \adjustbox{raise=0.5em}{%
                        $\eqnmarkbox[BrickRed]{dep1}{\strut We}$
                        \hspace{-0.3em}
                        $\eqnmarkbox[WildStrawberry]{dep2}{\strut evaluated}$
                        \hspace{-0.3em}
                        $\eqnmarkbox[Plum]{dep3}{\strut 11}$
                        \hspace{-0.3em}
                        $\eqnmarkbox[NavyBlue]{dep5}{\strut light-weight}$
                        \hspace{-0.3em}
                        $\eqnmarkbox[Cyan]{dep6}{\strut open-source}$
                        \hspace{-0.3em}
                        $\eqnmarkbox[OliveGreen]{dep4}{\strut ViLLMs}$
                    }
                \end{equation*}

                \annotatetwo[ yshift=0.8em]{above}{dep2}{dep1}{\scriptsize \itshape Subject}
                \annotatetwo[ yshift=0.8em]{above}{dep2}{dep4}{\scriptsize \itshape Direct Object}
                \annotatetwo[xshift=1.94cm, yshift=-0.8em]{below}{dep4}{dep3}{\scriptsize \itshape Number Modifier}
                \annotatetwo[xshift=0.8cm, yshift=2em]{above}{dep4}{dep5}{\scriptsize \itshape Adjective Modifier}
                \annotatetwo[yshift=-0.3em]{below}{dep5}{dep6}{\scriptsize \itshape Compound Modifier}
            \end{minipage}
        }
    }
    \vspace*{-0.1cm}
\caption{Illustration of Dependency Length computation with annotated syntactic relations.  
Arrows represent hierarchical dependencies between words, with colors corresponding to their head words for clarity. Then, the dependency length of the sentence is calculated as \(DL = \frac{1+4+3+2+1}{5}\).}
    \label{fig:dependencylength}
\end{figure}

\medskip
\noindent
\textbf{Syntactic Analysis} \quad  
Structural analysis is a key component of our novel LLM evaluation framework, revealing syntactic inefficiencies that traditional metrics often overlook. We introduce five distinctive evaluation metrics: \textit{(i) Word Count}, which analyzes meaningful word segments in an answer; \textit{(ii) Part-of-Speech Ratio} (POS Ratio), which measures information richness by computing the ratio of content to function words in the LLM's response, as illustrated in Figure \ref{fig:postagging}; \textit{(iii) Phrase Ratio}, similar to the POS Ratio but at the phrase level, as shown in Figure \ref{fig:phrasetagging}; \textit{(iv) Named Entity Difference} (NE Diff.), which quantifies discrepancies in named entities (e.g., people, locations) between the answer and question to assess content consistency; and \textit{(v) Dependency Length} (Dep. Length), which reflects sentence complexity by averaging the distance between heads and their dependents over the total dependencies, as illustrated in Figure \ref{fig:dependencylength}. For implementation, we employ two renowned \textit{Vietnamese Natural Language Processing toolkits}, namely \textit{Stanza} \cite{qi2020stanza} for POS tagging and \textit{underthesea}\footnote{\url{https://github.com/undertheseanlp/underthesea}} for other tasks contributing to our evaluation method.

\vspace*{-0.1cm}
\vspace*{-0.5cm}

\vspace*{-0.3cm}
\noindent
\textbf{Benchmarking Results} \quad  
To rank ViLLMs across different question types in CSConDa and determine the overall ranking, let \( r_{i,T} \) be the rank of model \( X \) for metric \( i \) within type \( T \), and \( R_T(X) \) denote its ranking for \( T \). The performance score for each type and the overall ranking are computed as Equation \ref{eq:ranking}.
\begin{equation} \label{eq:ranking}
\text{Score}_{X,T} = \frac{1}{\sum_{i} r_{i,T}}, \quad
\text{Score}_{X,\text{overall}} = \frac{1}{\sum_{T} R_T(X)}.
\end{equation}
A higher score indicates better performance. This ranking is derived solely from automatic metrics, while syntactic analysis provides complementary insights beyond quantitative evaluation. 


\subsection{Benchmarking Results}

\begin{figure}[!ht]
\vspace*{-1.1cm}
    \centering
    \begin{subfigure}{0.193\textwidth}
        \centering
        \includegraphics[width=\linewidth]{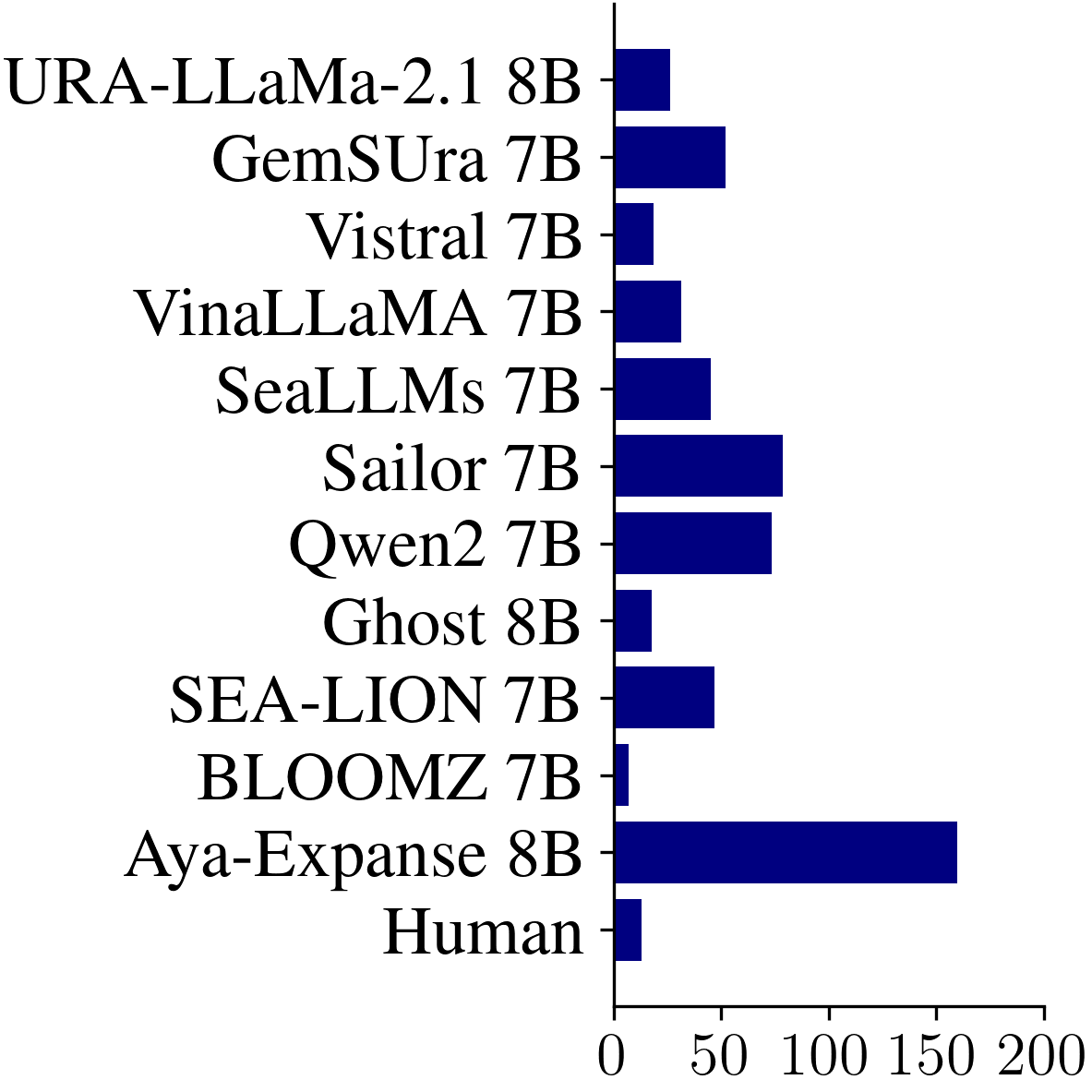}
        \caption*{\scriptsize Word Count}
    \end{subfigure}
    \begin{subfigure}{0.193\textwidth}
        \centering
        \includegraphics[width=\linewidth]{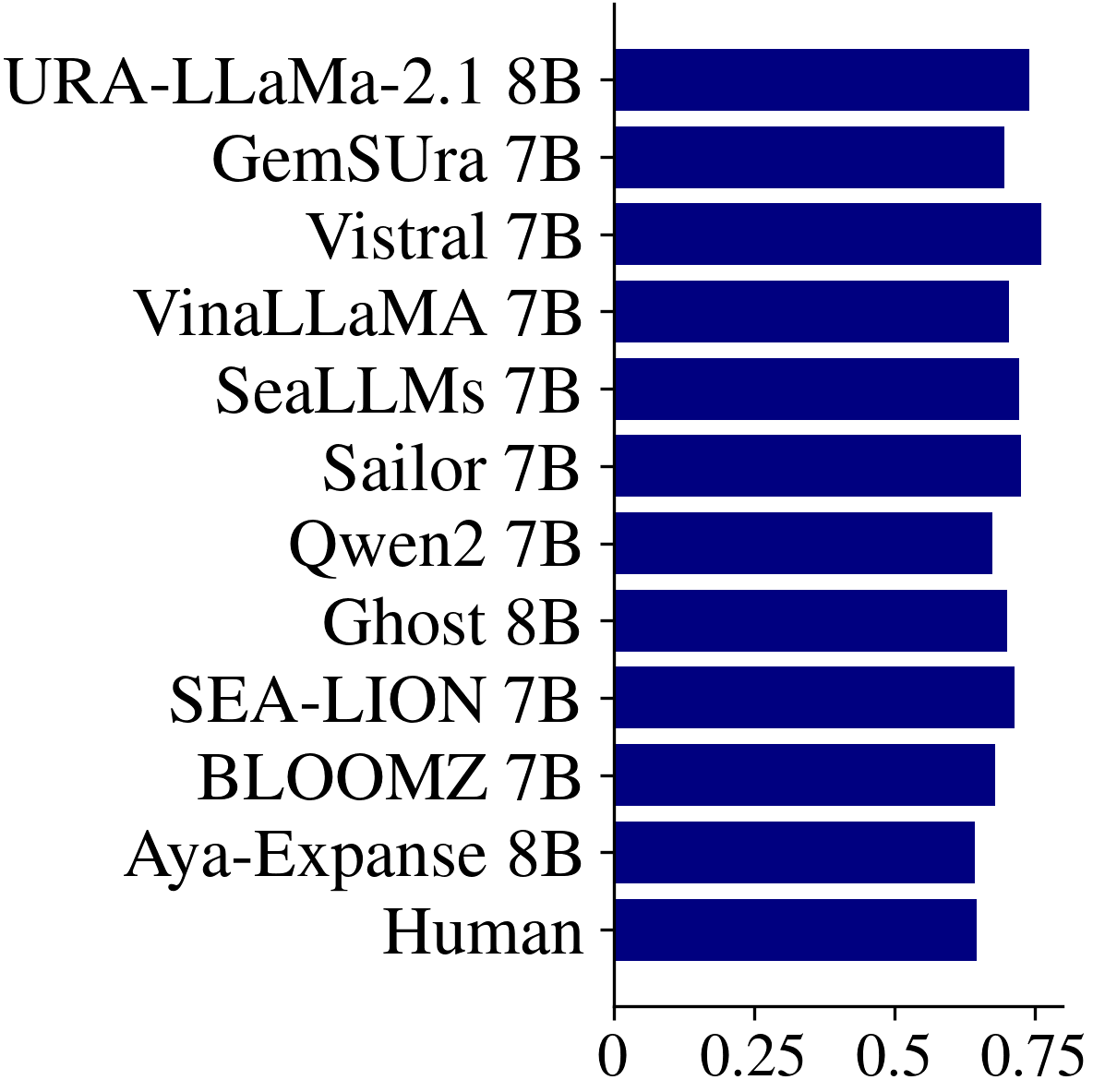}
        \caption*{\scriptsize POS Ratio}
    \end{subfigure}
    \begin{subfigure}{0.193\textwidth}
        \centering
        \includegraphics[width=\linewidth]{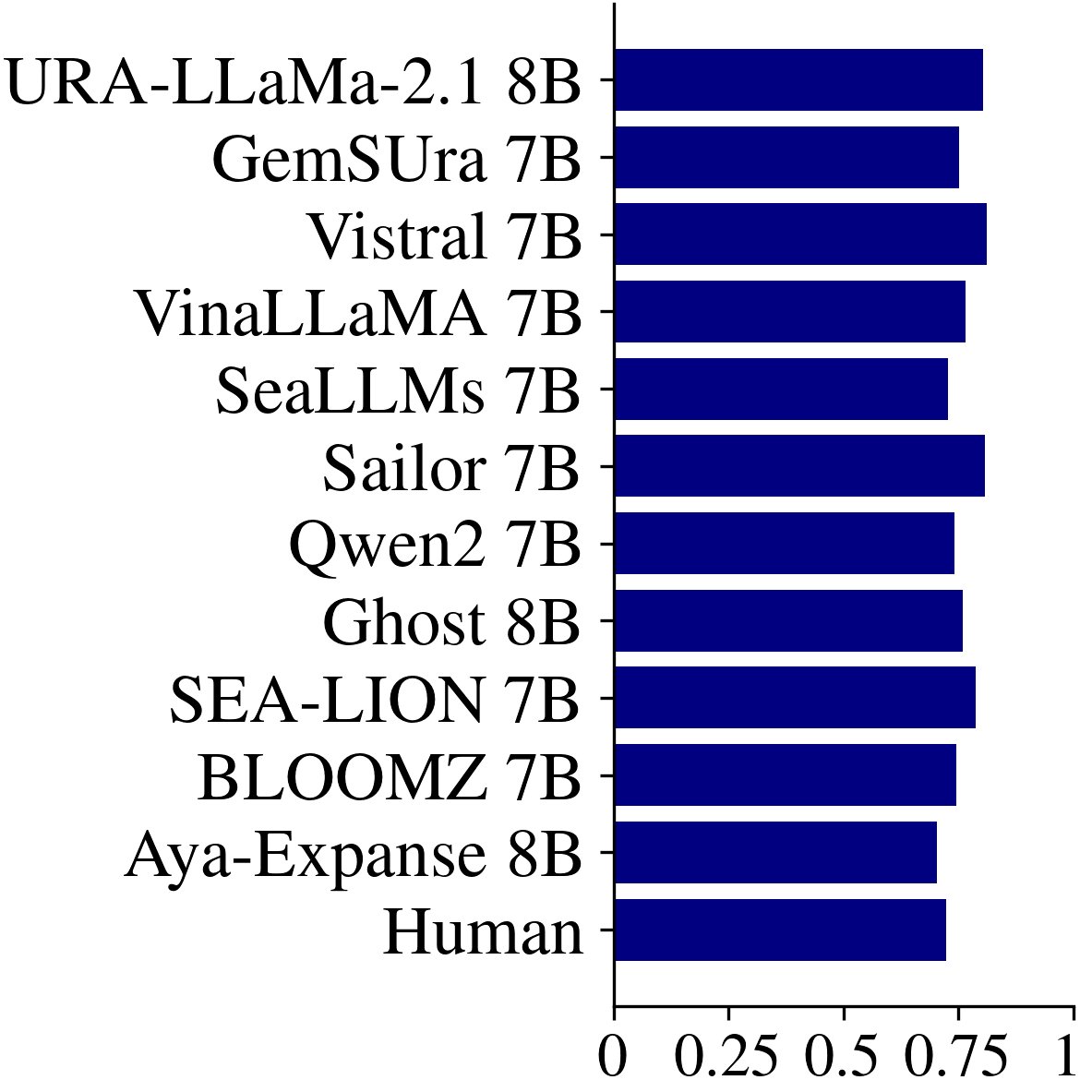}
        \caption*{\scriptsize Phrase Ratio}
    \end{subfigure}
    \begin{subfigure}{0.193\textwidth}
        \centering
        \includegraphics[width=\linewidth]{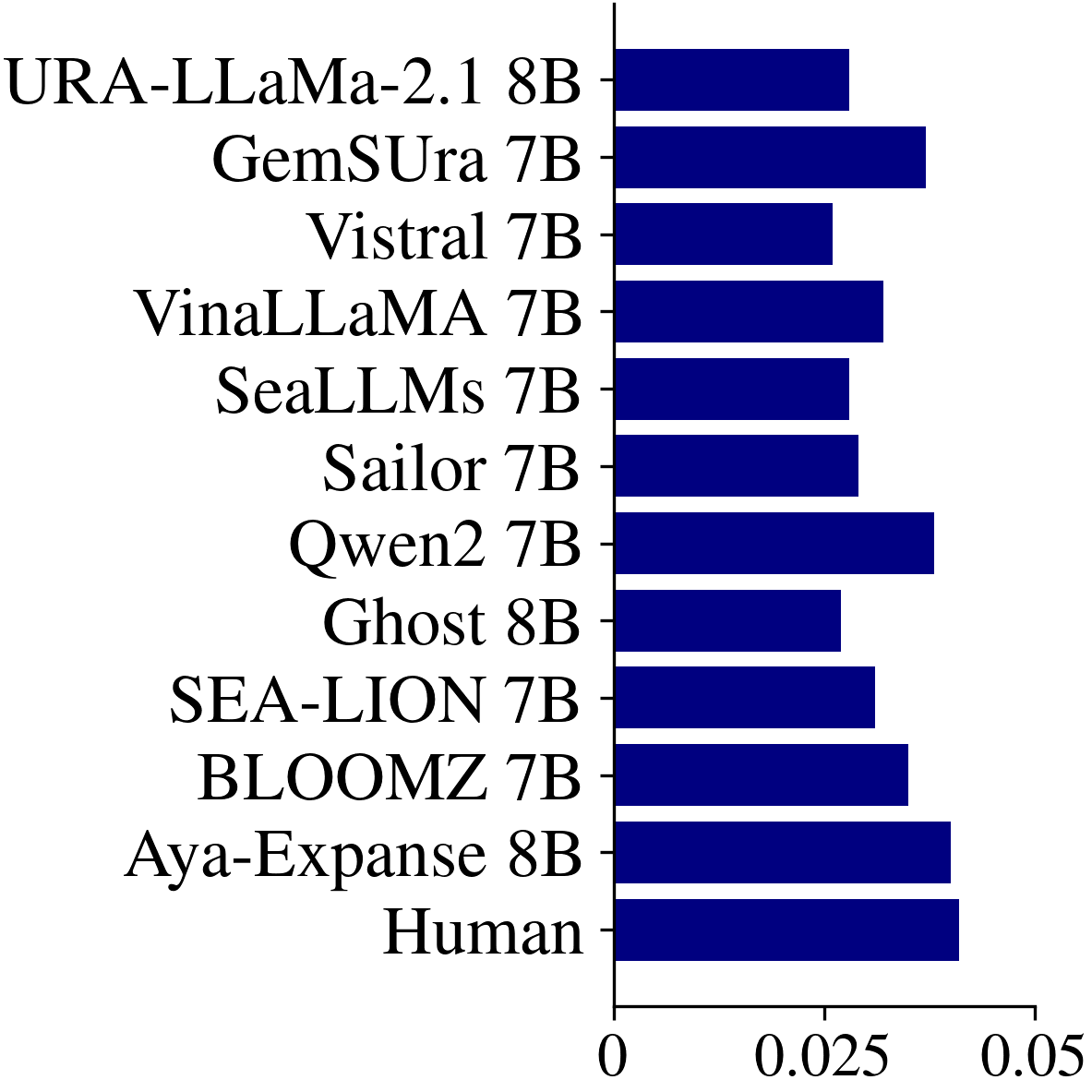}
        \caption*{\scriptsize NE Diff.} 
    \end{subfigure}
    \begin{subfigure}{0.193\textwidth}
        \centering
        \includegraphics[width=\linewidth]{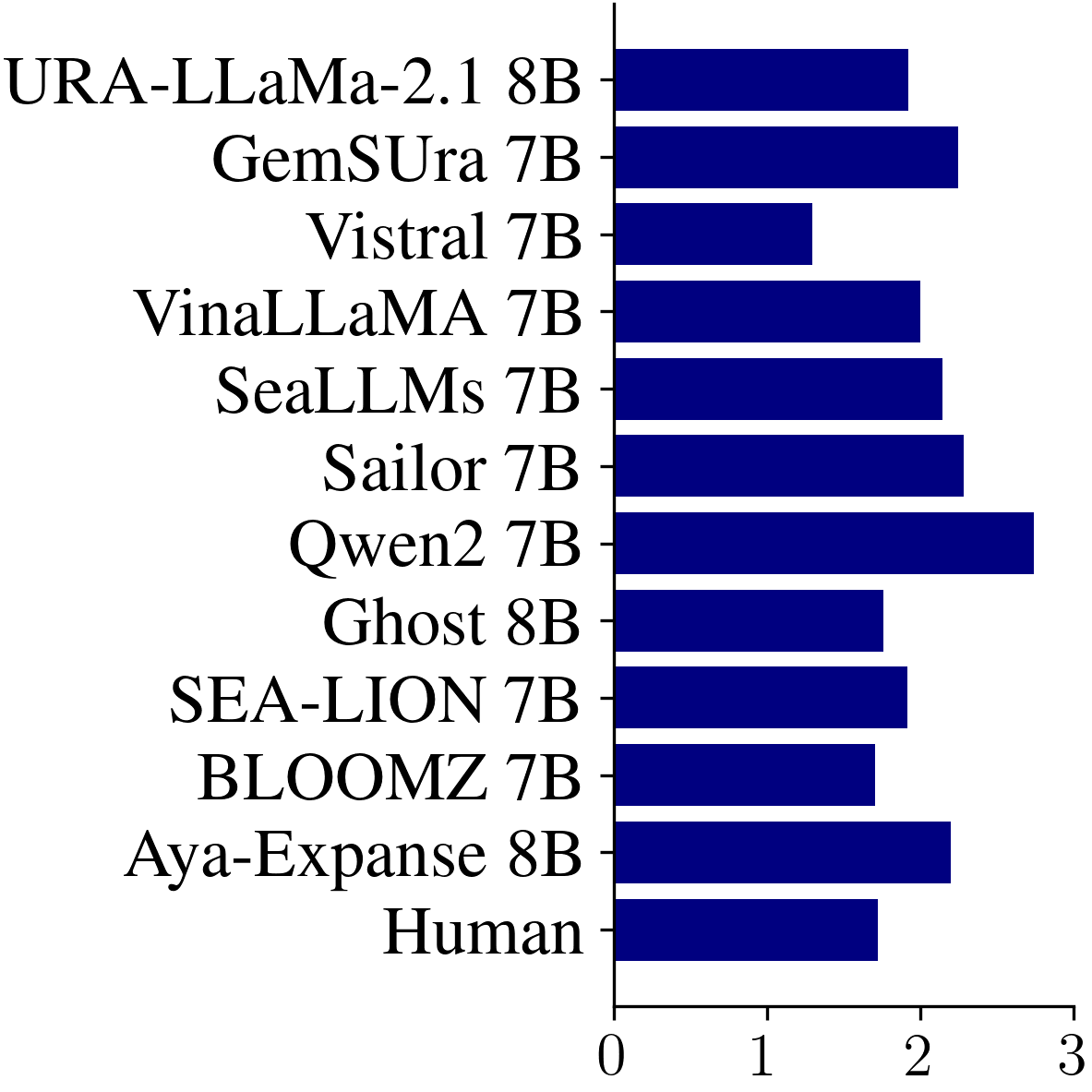}
        \caption*{\scriptsize Dep. Length}
    \end{subfigure}
    \vspace*{-0.5cm}
    \caption{Syntactic analysis of general-type questions.}
    \label{fig:syngeneral}
\end{figure}

\begin{figure}[!ht]
\vspace*{-0.9cm}
    \centering
    \begin{subfigure}{0.193\textwidth}
        \centering
        \includegraphics[width=\linewidth]{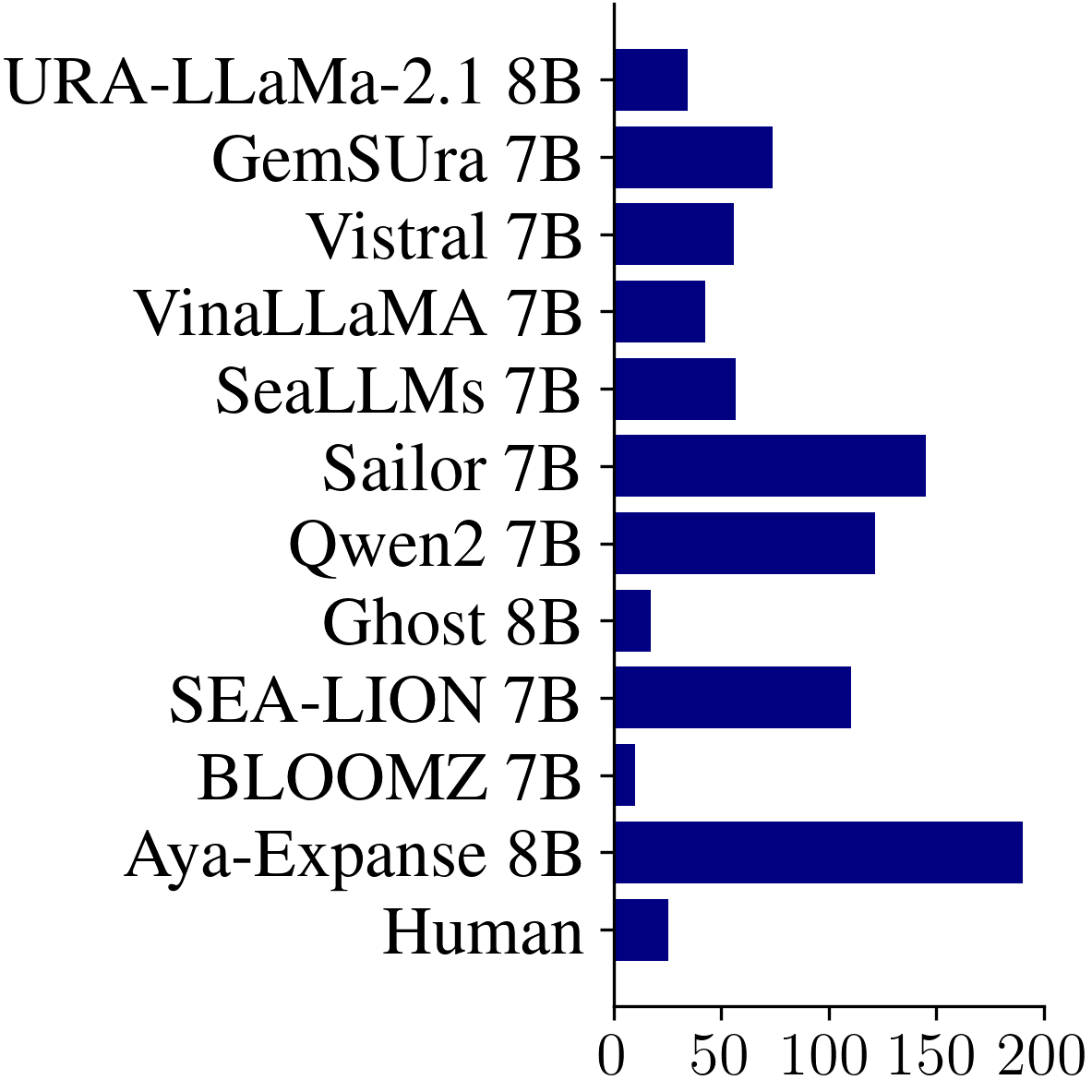}
        \caption*{\scriptsize Word Count}
    \end{subfigure}
    \begin{subfigure}{0.193\textwidth}
        \centering
        \includegraphics[width=\linewidth]{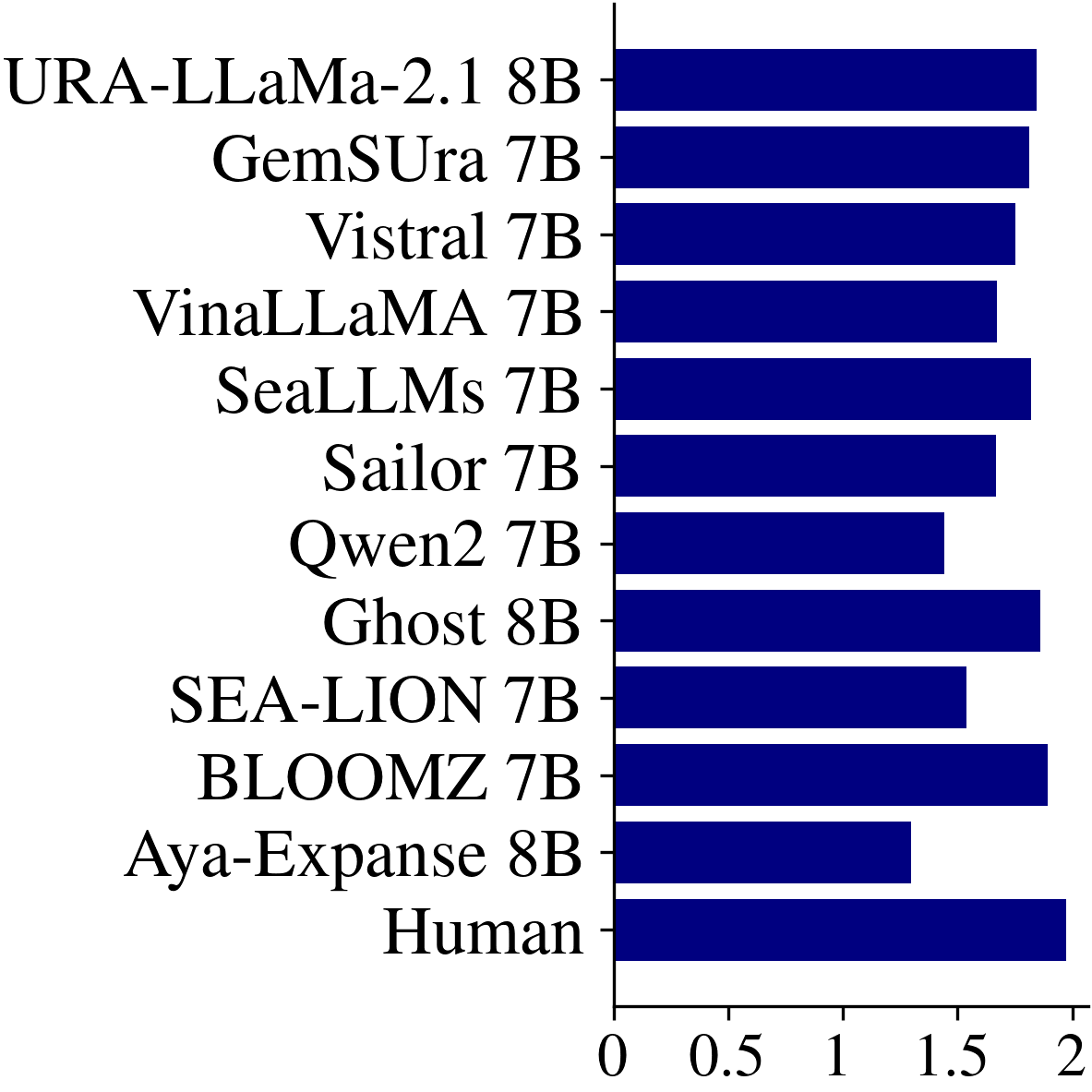}
        \caption*{\scriptsize POS Ratio}
    \end{subfigure}
    \begin{subfigure}{0.193\textwidth}
        \centering
        \includegraphics[width=\linewidth]{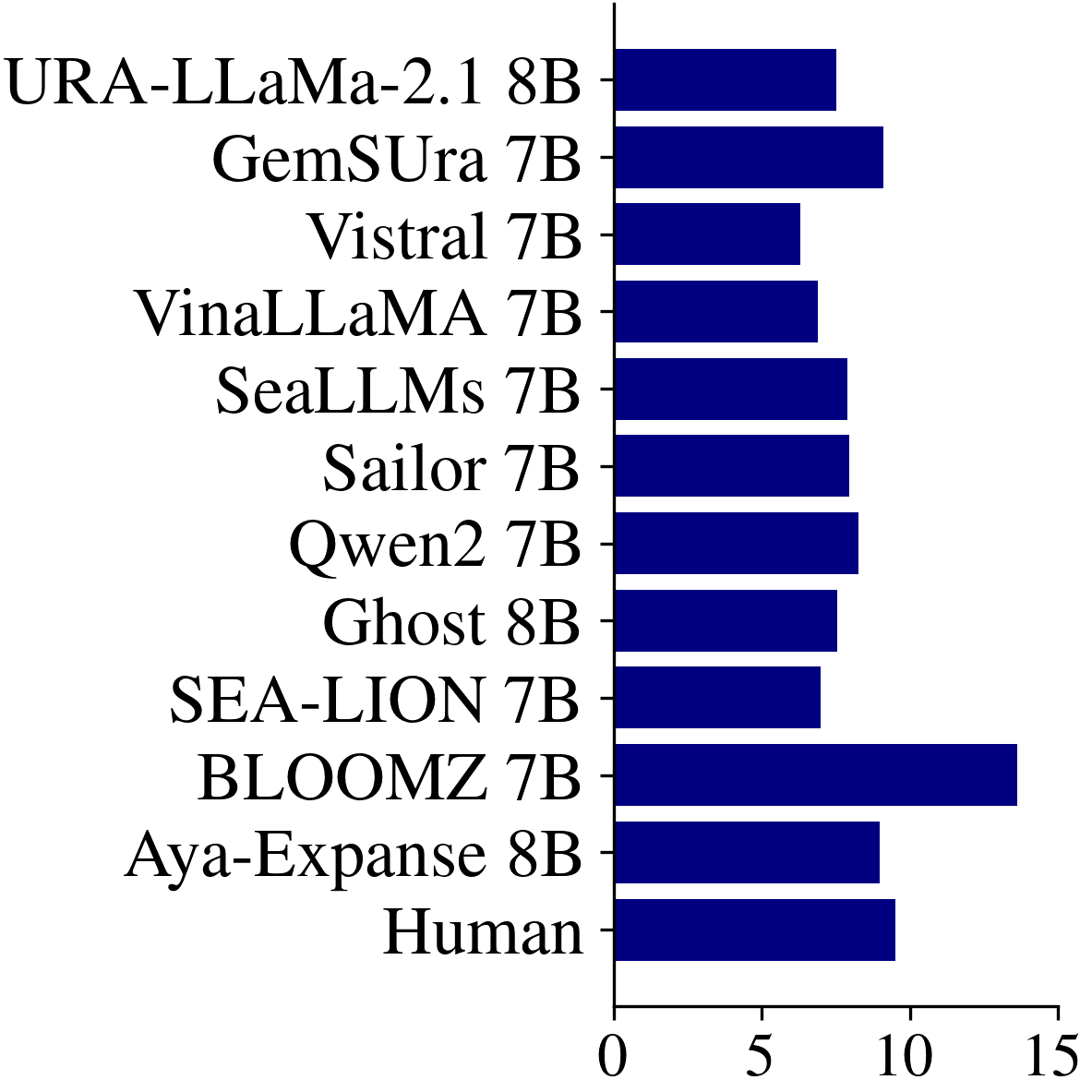}
        \caption*{\scriptsize Phrase Ratio}
    \end{subfigure}
    \begin{subfigure}{0.193\textwidth}
        \centering
        \includegraphics[width=\linewidth]{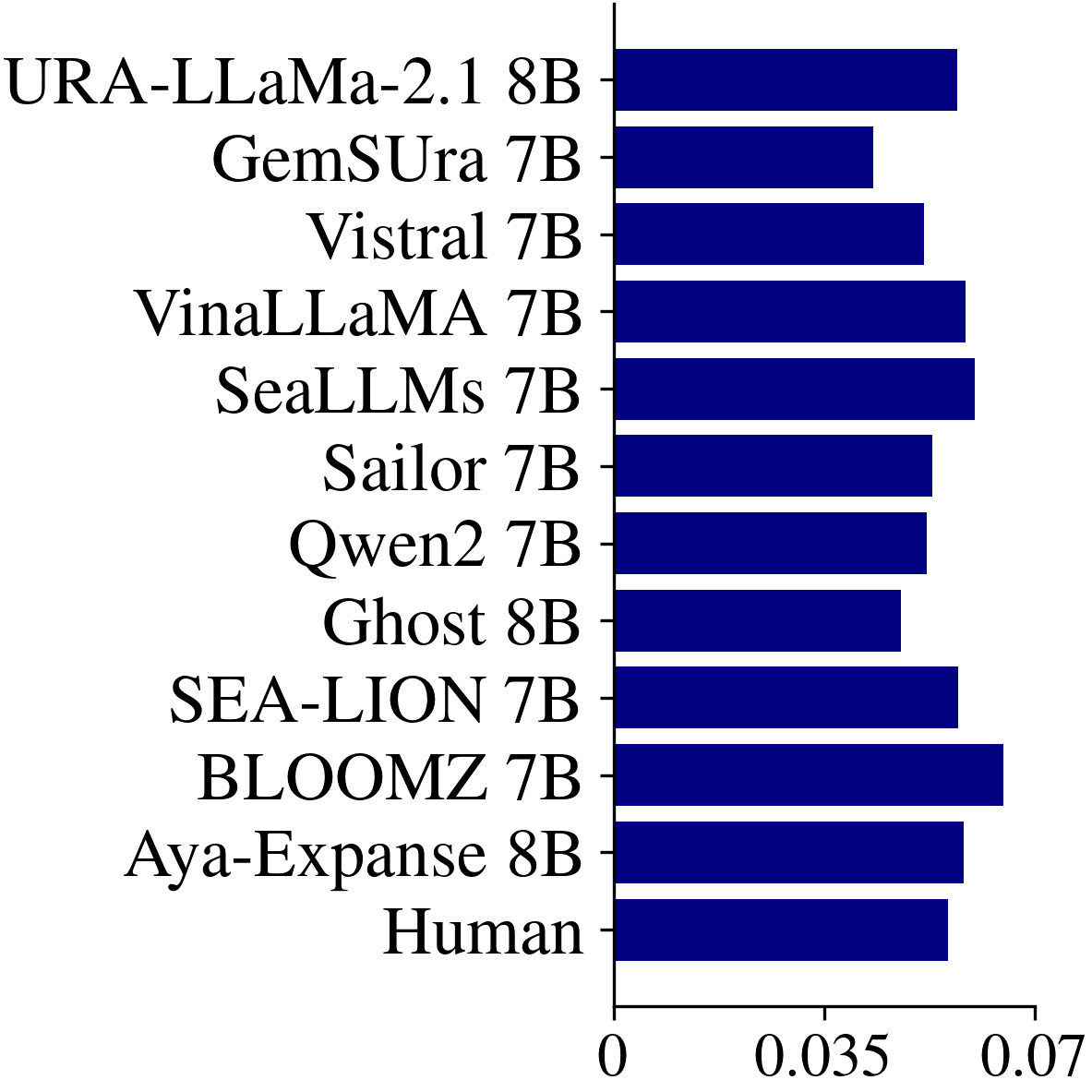}
        \caption*{\scriptsize NE Diff.} 
    \end{subfigure}
    \begin{subfigure}{0.193\textwidth}
        \centering
        \includegraphics[width=\linewidth]{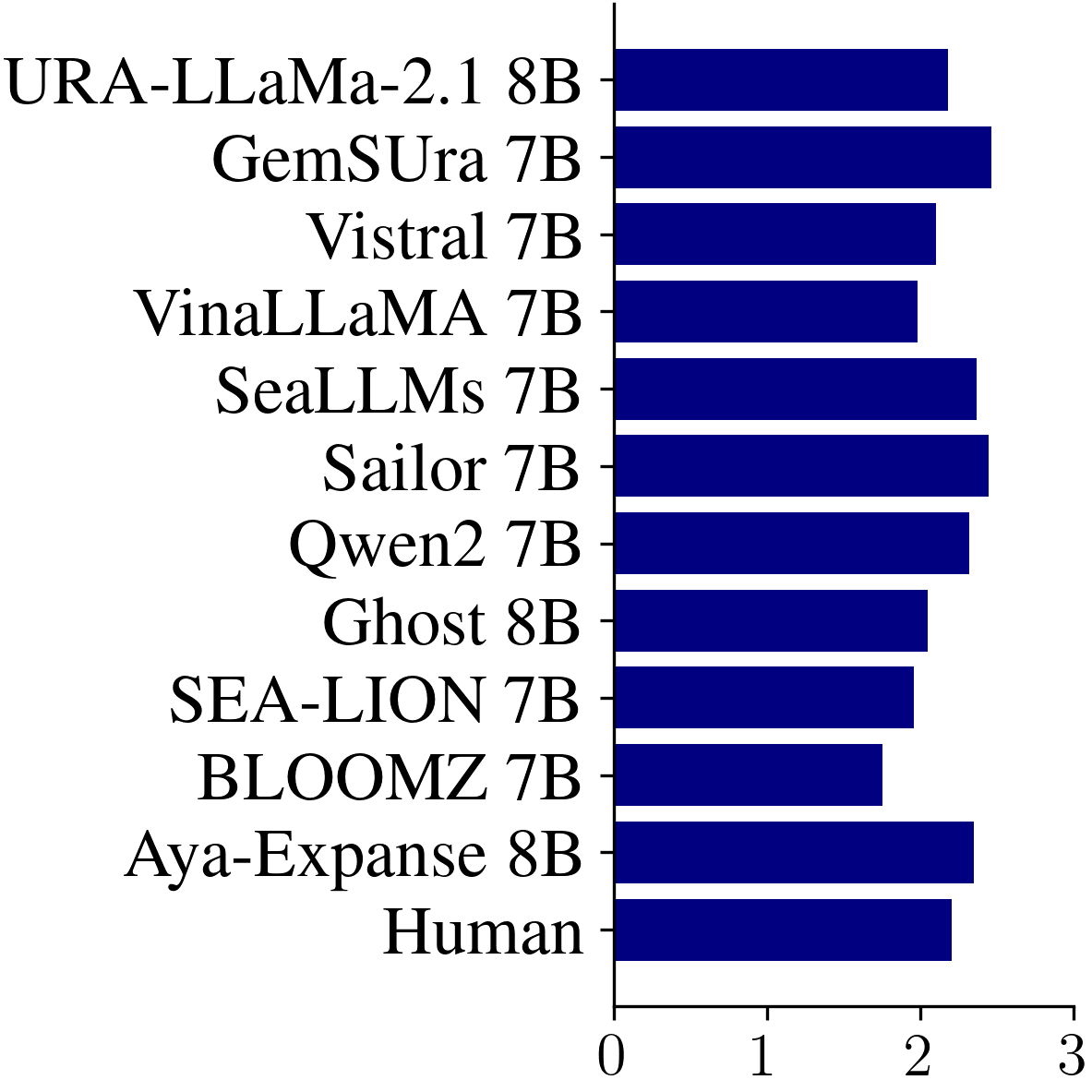}
        \caption*{\scriptsize Dep. Length}
    \end{subfigure}
        \vspace*{-0.5cm}
    \caption{Syntactic analysis of simple-type questions.}
    \label{fig:synsimple}
\end{figure}

\begin{figure}[!ht]
\vspace*{-0.9cm}
    \centering
    \begin{subfigure}{0.193\textwidth}
        \centering
        \includegraphics[width=\linewidth]{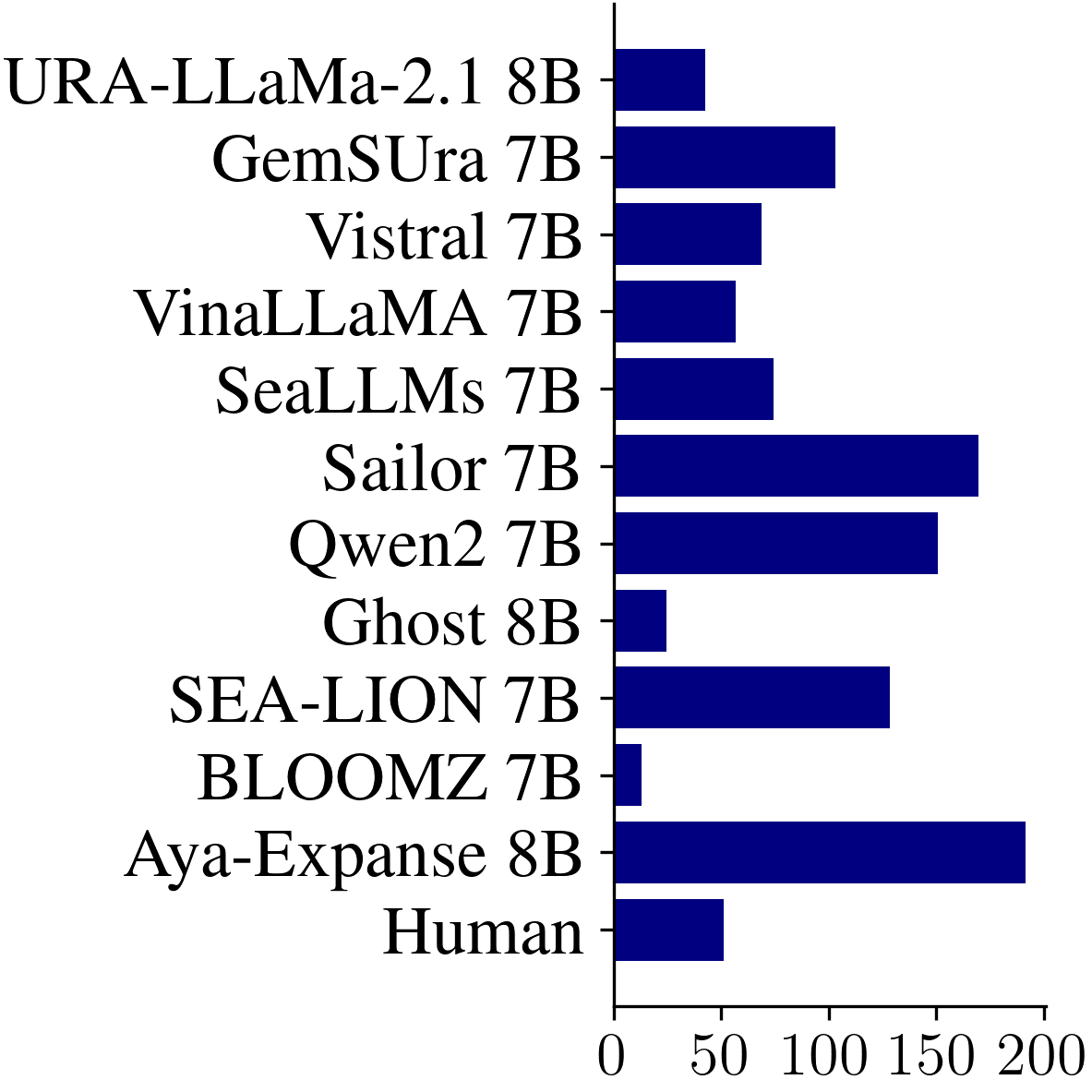}
        \caption*{\scriptsize Word Count}
    \end{subfigure}
    \begin{subfigure}{0.193\textwidth}
        \centering
        \includegraphics[width=\linewidth]{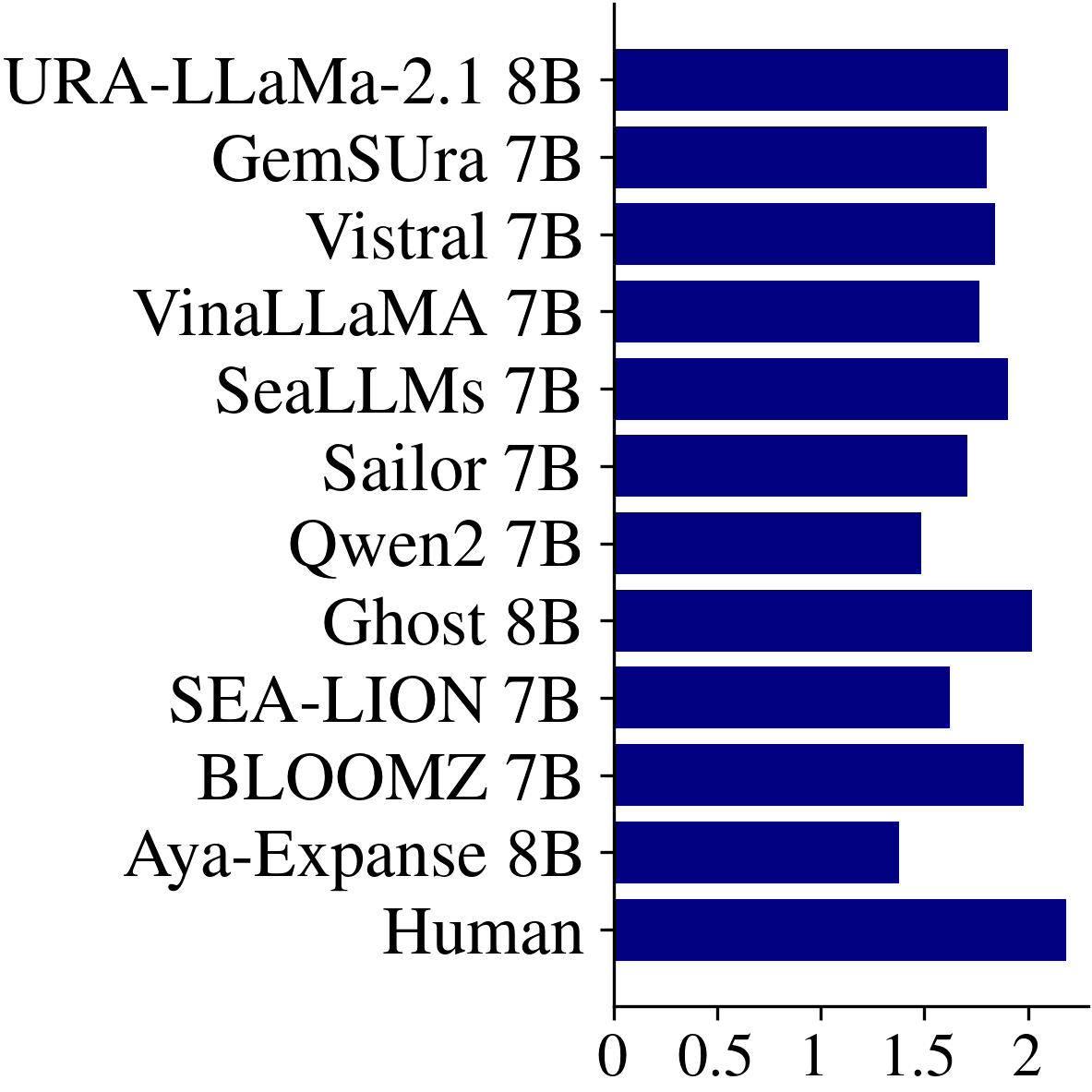}
        \caption*{\scriptsize POS Ratio}
    \end{subfigure}
    \begin{subfigure}{0.193\textwidth}
        \centering
        \includegraphics[width=\linewidth]{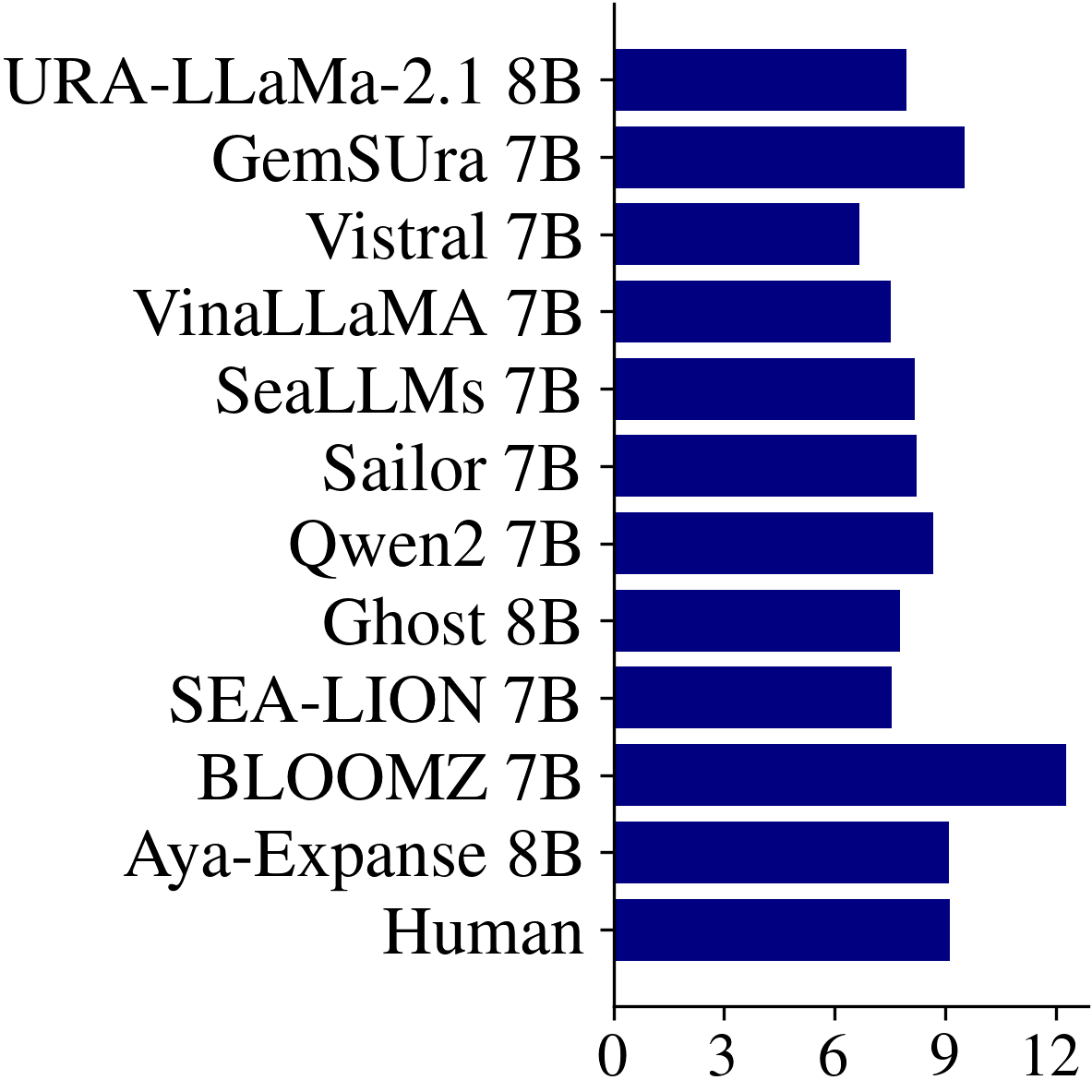}
        \caption*{\scriptsize Phrase Ratio}
    \end{subfigure}
    \begin{subfigure}{0.193\textwidth}
        \centering
        \includegraphics[width=\linewidth]{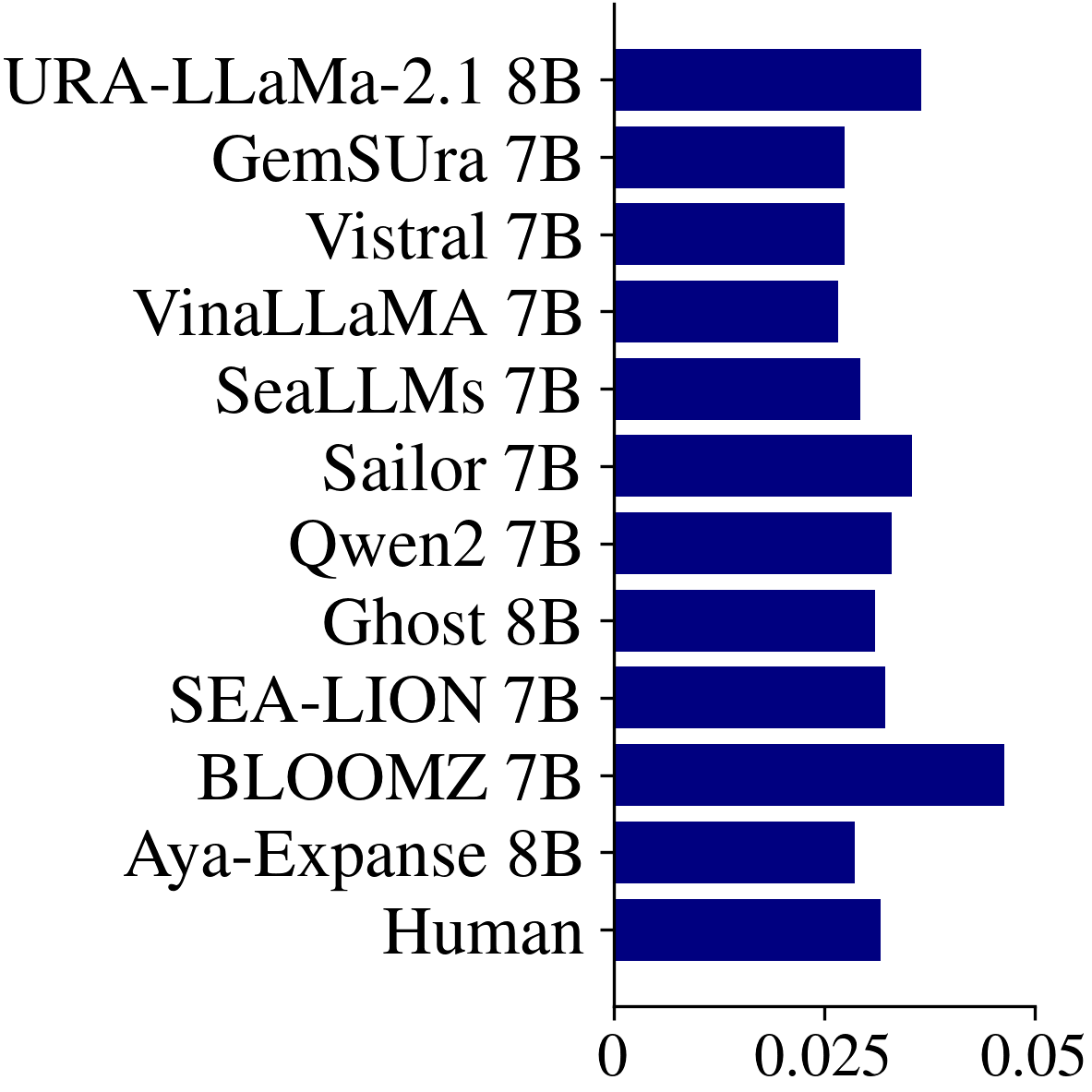}
        \caption*{\scriptsize NE Diff.} 
    \end{subfigure}
    \begin{subfigure}{0.193\textwidth}
        \centering
        \includegraphics[width=\linewidth]{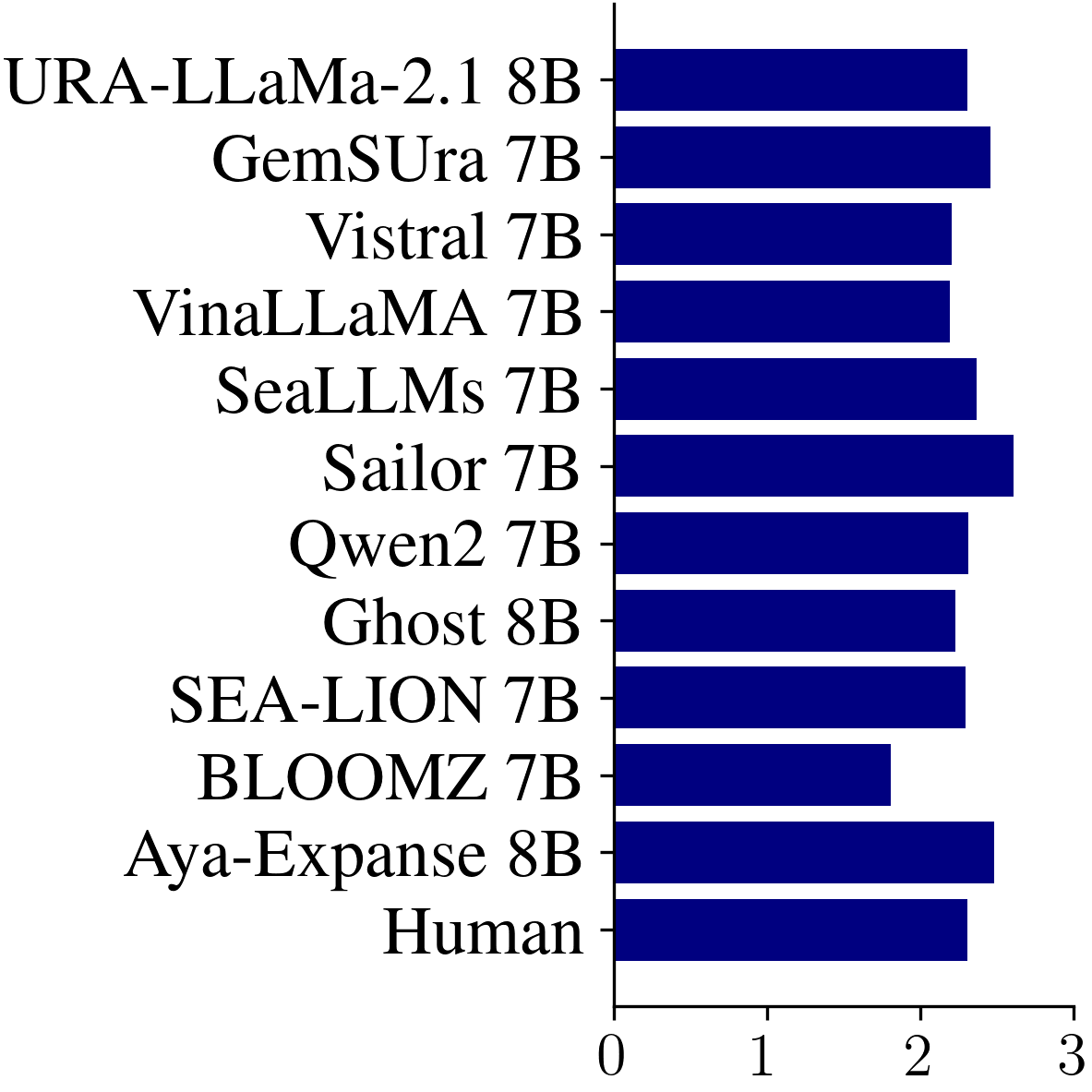}
        \caption*{\scriptsize Dep. Length}
    \end{subfigure}
        \vspace*{-0.5cm}
    \caption{Syntactic analysis of complex-type questions.}
    \label{fig:syncomplex}
\end{figure}

Table \ref{tab:general}, Table \ref{tab:simple}, and Table \ref{tab:complex} present benchmark scores of selected ViLLMs across multiple evaluation metrics for the three CSConDa categories, providing a comparative view of model performance. Additionally, Figure \ref{fig:syngeneral}, Figure \ref{fig:synsimple}, and Figure \ref{fig:syncomplex} illustrate key syntactic patterns, offering insights into how ViLLMs process different linguistic structures and their impact on performance variations.






\subsection{Analysis and Insights}

We analyze the benchmarking results and highlight the following key insights.

\medskip \noindent \textbf{Performance of Lightweight Open-Source ViLLMs in Customer Support.} \quad
Based on the overall ranking, the top five models are Vistral 7B, which achieves the strongest results, followed by SeaLLMs 7B, Sailor 7B, SEA-LION 7B, and GemSUra 7B. However, while these models lead among those evaluated, they are not definitive solutions. Their poor performance on CSConDa benchmarks suggests that current ViLLMs still struggle to generate accurate and fluent responses, particularly when handling queries with linguistic variations.

\medskip \noindent \textbf{Vietnamese-Finetuned LLMs vs. Multilingual Models.} \quad
Our evaluation finds no significant performance gap between LLMs fine-tuned for Vietnamese and multilingual models supporting Vietnamese. This result is expected due to the limited availability of high-quality customer support data in public datasets.

\medskip \noindent \textbf{ViLLMs Lack Human-Like Writing Styles.} \quad
While BERTScore remains acceptable across query types, Cos. Sim. and other lexical-based metrics perform significantly worse. This suggests that ViLLMs produce semantically similar words with different surface forms, leading to responses that deviate from natural human-like writing in this domain.

\begin{table*}[!ht]
    \centering
    \caption{Benchmarking results of ViLLMs on general-type questions.}
    \renewcommand{\arraystretch}{1.1}
    \setlength{\tabcolsep}{3pt} 
    \resizebox{\textwidth}{!}{ 
    \begin{tabular}{lcccccc}
        \toprule
        \textbf{Model Name} & \textbf{BLEU-2} \(\uparrow\) & \textbf{ROUGE-L} \(\uparrow\) & \textbf{METEOR} \(\uparrow\) & \textbf{Cos. Sim.} \(\uparrow\) & \textbf{BERTScore} \(\uparrow\) & \textbf{Hallu. Score} \(\downarrow\) \\
        \midrule
        URA-LLaMa-2.1 8B & 0.004 & \textbf{0.161} & 0.072 & 0.219 & \textbf{0.654} & 0.782 \\
        GemSUra 7B & 0.006 & 0.151 & \textbf{0.087} & 0.226 & 0.645 & 0.694 \\
        Vistral 7B & 0.005 & \textbf{0.171} & \textbf{0.089} & \textbf{0.269} & \textbf{0.659} & \textbf{0.522} \\
        VinaLLaMA 7B & \textbf{0.008} & 0.153 & 0.087 & \textbf{0.251} & 0.645 & \textbf{0.518} \\
        \hdashline
        SeaLLMs 7B & 0.001 & \textbf{0.166} & 0.047 & 0.247 & \textbf{0.661} & 0.598 \\
        Sailor 7B & \textbf{0.012} & \textbf{0.165} & 0.080 & \textbf{0.246} & \textbf{0.662} & \textbf{0.588} \\
        Qwen2 7B & 0.006 & 0.133 & \textbf{0.086} & \textbf{0.261} & 0.634 & 0.612 \\
        Ghost 8B & \textbf{0.009} & 0.142 & \textbf{0.091} & 0.238 & 0.632 & 0.664 \\
        SEA-LION 7B & \textbf{0.011} & \textbf{0.188} & 0.086 & 0.241 & \textbf{0.678} & \textbf{0.452} \\
        BLOOMZ 7B & 0.003 & 0.137 & 0.046 & 0.223 & 0.634 & \textbf{0.448} \\ 
        Aya-Expanse 8B & \textbf{0.010} & 0.102 & \textbf{0.086} & \textbf{0.260} & 0.601 & 0.798 \\
        \bottomrule
    \end{tabular}
    }
    \label{tab:general}
\end{table*}

\vspace*{-0.7cm}
\begin{table*}[ht]
    \centering
    \caption{Benchmarking results of ViLLMs on simple-type questions.}
    \renewcommand{\arraystretch}{1.1}
    \setlength{\tabcolsep}{3pt} 
    \resizebox{\textwidth}{!}{ 
    \begin{tabular}{lcccccc}
        \toprule
        \textbf{Model Name} & \textbf{BLEU-2} \(\uparrow\) & \textbf{ROUGE-L} \(\uparrow\) & \textbf{METEOR} \(\uparrow\) & \textbf{Cos. Sim.} \(\uparrow\) & \textbf{BERTScore} \(\uparrow\) & \textbf{Hallu. Score} \(\downarrow\) \\
        \midrule
        URA-LLaMa-2.1 8B & 0.009 & \textbf{0.221} & 0.075 & 0.266 & \textbf{0.664} & 0.878 \\
        GemSUra 7B & \textbf{0.015} & \textbf{0.213} & \textbf{0.104} & 0.274 & 0.659 & 0.846 \\
        Vistral 7B & \textbf{0.016} & \textbf{0.227} & \textbf{0.104} & \textbf{0.325} & \textbf{0.675} & \textbf{0.746} \\
        VinaLLaMA 7B & 0.011 & 0.191 & 0.102 & 0.322 & 0.648 & 0.868 \\
        \hdashline
        SeaLLMs 7B & \textbf{0.015} & \textbf{0.218} & \textbf{0.105} & \textbf{0.329} & \textbf{0.667} & \textbf{0.568} \\
        Sailor 7B & 0.012 & \textbf{0.213} & 0.088 & 0.304 & \textbf{0.670} & \textbf{0.774} \\
        Qwen2 7B & 0.010 & 0.173 & \textbf{0.112} & \textbf{0.337} & 0.640 & 0.894 \\
        Ghost 8B & 0.012 & 0.187 & \textbf{0.110} & \textbf{0.323} & 0.645 & 0.900 \\
        SEA-LION 7B & \textbf{0.013} & 0.206 & 0.073 & 0.253 & \textbf{0.674} & \textbf{0.616} \\
        BLOOMZ 7B & 0.003 & 0.159 & 0.039 & 0.168 & 0.635 & \textbf{0.674} \\
        Aya-Expanse 8B & \textbf{0.017} & 0.171 & \textbf{0.115} & \textbf{0.347} & 0.631 & 0.984 \\
        \bottomrule
    \end{tabular}}
    \label{tab:simple}
\end{table*}
\pagebreak
\begin{table*}[!ht]
    \centering
   \caption{Benchmarking results of ViLLMs on complex-type questions.}
    \renewcommand{\arraystretch}{1.1}
    \setlength{\tabcolsep}{3pt} 
    \resizebox{\textwidth}{!}{ 
    \begin{tabular}{lcccccc}
        \toprule
        \textbf{Model Name} & \textbf{BLEU-2} \(\uparrow\) & \textbf{ROUGE-L} \(\uparrow\) & \textbf{METEOR} \(\uparrow\) & \textbf{Cos. Sim.} \(\uparrow\) & \textbf{BERTScore} \(\uparrow\) & \textbf{Hallu. Score} \(\downarrow\) \\
        \midrule
        URA-LLaMa-2.1 8B & 0.011 & \textbf{0.238} & 0.075 & 0.298 & \textbf{0.659} & 0.918 \\
        GemSUra 7B & \textbf{0.022} & \textbf{0.242} & \textbf{0.107} & 0.335 & 0.658 & 0.902 \\
        Vistral 7B & \textbf{0.019} & \textbf{0.247} & \textbf{0.098} & 0.349 & \textbf{0.667} & \textbf{0.782} \\
        VinaLLaMA 7B & \textbf{0.019} & \textbf{0.232} & \textbf{0.107} & \textbf{0.356} & 0.652 & 0.912 \\
        \hdashline
        SeaLLMs 7B & \textbf{0.022} & \textbf{0.247} & 0.105 & \textbf{0.365} & \textbf{0.666} & \textbf{0.642} \\
        Sailor 7B & 0.015 & 0.231 & 0.089 & 0.345 & \textbf{0.665} & \textbf{0.864} \\
        Qwen2 7B & 0.015 & 0.213 & \textbf{0.108} & \textbf{0.371} & 0.641 & 0.966 \\
        Ghost 8B & \textbf{0.020} & 0.228 & \textbf{0.108} & \textbf{0.361} & 0.648 & 0.932 \\
        SEA-LION 7B & 0.012 & 0.217 & 0.067 & 0.288 & \textbf{0.670} & \textbf{0.662} \\
        BLOOMZ 7B & 0.004 & 0.146 & 0.034 & 0.168 & 0.606 & \textbf{0.712} \\
        Aya-Expanse 8B & \textbf{0.027} & 0.223 & \textbf{0.115} & \textbf{0.393} & 0.641 & 0.982 \\
        \bottomrule
    \end{tabular}}
    \label{tab:complex}
\end{table*}

\medskip \noindent \textbf{ViLLMs Are Verbose and Structurally Inflated.} \quad
Syntactic analysis shows that human responses are consistently more concise across all query types, as indicated by their lower word count. In contrast, ViLLMs over-generate words and sentences, producing longer responses. Furthermore, Hallu. Score is lowest for general-type questions, where ViLLMs provide the shortest answers. These findings suggest that verbosity induces semantic drift, reducing alignment with the original query and increasing the likelihood of errors and hallucinations.

\medskip \noindent \textbf{ViLLMs Exhibit Stronger Structural Dependency.} \quad
The Dep. Length metric shows that human responses consistently have the shortest dependency lengths, indicating minimal syntactic complexity and greater structural flexibility. This suggests a more fluid and unstructured style, characteristic of human-like responses. In contrast, ViLLMs, trained predominantly on structured data, exhibit higher dependency parsing scores, leading to rigid sentence constructions and a writing style that lacks natural variation.

\medskip \noindent \textbf{Towards Structurally-Aware Fine-Tuning of ViLLMs.} \quad
Although POS Ratio, Phrase Ratio, and NE Diff. metrics indicate that ViLLMs maintain internal consistency, their responses remain overly long and structurally rigid, often leading to semantic drift and increased errors. Our comprehensive benchmarking highlights structural constraints as a key barrier to semantic quality, particularly in customer support interactions, where natural and adaptive communication is essential. Addressing these limitations requires a refined fine-tuning approach that enhances structural efficiency, promotes syntactic adaptability, and optimizes linguistic economy.

\section{Conclusion, Limitation, and Future Work}

In this work, we introduce CSConDa, the first Vietnamese QA dataset for benchmarking customer support interactions, alongside a novel evaluation framework. Our approach extends beyond conventional lexical and semantic assessments by integrating advanced hallucination detection and syntactic analysis. We outline the dataset construction process and provide an in-depth linguistic characterization, highlighting CSConDa’s distinctions from existing resources. Using the test split, we conduct a comprehensive evaluation of SOTA lightweight open-source ViLLMs, following a survey underscoring the need for rigorous benchmarking. Our findings indicate that while LLMs demonstrate strong grammatical accuracy and fluency, they struggle with structural efficiency and adaptability. In contrast, human responses prioritize brevity, clarity, and contextual relevance, reinforcing the importance of structurally-aware fine-tuning to bridge this gap. CSConDa, along with our evaluation results, offers enterprises a valuable tool for identifying suitable ViLLMs and guiding the development of effective QA systems. Despite the robustness of our evaluation framework, its primary limitation is focusing solely on intrinsic model capabilities. Future work could enhance CSConDa by incorporating contextual information, including relevant details essential for answering queries, ensuring such extensions preserve its realism and task-oriented nature for practical applications. 

\bibliographystyle{ieeetr}
\bibliography{references}

\end{document}